# Large Language Models Predict Functional Outcomes after Acute Ischemic Stroke

Anjali K. Kapoor, A.B.[1,*], Anton Alyakin, M.S.E[1-3], Jin Vivian Lee, M.D. M.S.,[1-3], Eunice Yang, B.S.[1,4], Annelene M. Schulze, M.S.[1], Krithik Vishwanath[1,5], Jinseok Lee, Ph.D.[2,6], Yindalon Aphinyanaphongs, M.D., Ph.D.[7,8], Howard Riina, M.D.[1,9], Jennifer A. Frontera, M.D.[10], Eric Karl Oermann, M.D.[1,2,8,11,‡]

[1]Department of Neurosurgery, NYU Langone Health; New York, NY 10016, USA
[2]Global AI Frontier Lab, New York University; Brooklyn, NY 11201, USA
[3]Department of Neurosurgery, Washington University in Saint Louis; Saint Louis, MO 63110, USA
[4]Columbia University Vagelos College of Physicians and Surgeons; New York, NY, USA
[5]Department of Aerospace Engineering & Engineering Mechanics, The University of Texas at Austin; Austin, TX 78705, USA
[6]Department of Biomedical Engineering, Kyung Hee University, Yongin, South Korea
[7]Department of Population Health, NYU Langone Health, New York, NY 10016, USA
[8]Division of Applied AI Technologies, NYU Langone Health, New York, NY 10016, USA
[9]Department of Radiology, NYU Langone Health; New York, NY 10016, USA
[10]Department of Neurology, NYU Langone Health; New York, NY 10016, USA
[11]Center for Data Science, New York University; New York, NY 10011, USA

***Corresponding author (review)**: Anjali K. Kapoor, Department of Neurosurgery, NYU Langone Medical Center, New York University, 550 First Ave, MS 3 205, New York, NY 10016, USA. Email: anjali.kapoor@nyulangone.org

[‡]**Corresponding author (post-publication)**: Eric K. Oermann, MD, Department of Neurosurgery, NYU Langone Medical Center, New York University, 550 First Ave, MS 3 205, New York, NY 10016, USA. Email: Eric.Oermann@nyulangone.org

**Author Contributions**: Conception and design: Frontera, Oermann. Registry maintenance: Frontera. Data acquisition: Kapoor, Frontera. Model training and evaluation: Kapoor. Baseline model reproduction: Kapoor. Analysis and interpretation of data: Kapoor, Alyakin, Oermann. Statistical analysis: Kapoor, Alyakin. Figure creation: Kapoor, Alyakin, Yang, Schulze, Oermann. Manuscript drafting: Kapoor, Alyakin, Lee, Oermann. Critical revision of manuscript: All authors. Study supervision: Frontera, Oermann. Resources and funding: Frontera, Oermann.

## Abstract

**Background:** Accurate prediction of functional outcomes measured after acute ischemic stroke can guide clinical decision-making and resource allocation. Prior work on predicting modified Rankin Scale has relied predominantly on structured variable predictors (age, NIHSS) and conventional machine learning. The ability of large language models (LLMs) to infer future mRS scores from routine clinical admission notes is largely unexplored.

**Methods:** We evaluated encoder (BERT, NYUTron) and generative (Llama-3.1-8B, MedGemma-4B) LLMs, in both frozen and fine-tuned settings, for discharge and 90-day mRS prediction using a large, real-world stroke registry. We retrospectively analyzed 11,472 admission History and Physical notes from 8,723 ischemic stroke admissions in the NYU Langone Get With The Guidelines–Stroke registry (2016–2025). The discharge outcome dataset included 9,485 notes, while the 90-day outcome dataset included 1,898 notes. Data were temporally split with the most recent 12 months held out for testing. Performance was assessed using exact (7-class) mRS accuracy and binary functional outcome (mRS 0–2 vs. 3–6) accuracy and compared against established structured-data baselines incorporating NIHSS and age.

**Results:** The best performing model was fine-tuned Llama, which achieved the highest 90-day exact mRS accuracy of 33.9% [95% CI, 27.9%-39.9%]. It also achieved 90-day binary functional outcome accuracy of 76.3% [95% CI, 70.7%-81.9%], discharge exact mRS accuracy of 42.0% [95% CI, 39.0%-45.0%], and discharge binary accuracy of 75.0% [95% CI, 72.4%-77.6%]. For 90-day prediction, Llama performed comparably to structured-data baselines (exact: 28.1%; binary: 72.6–73.8%).

**Conclusions:** Fine-tuned LLMs can predict post-stroke functional outcomes from admission notes alone, achieving performance comparable to models requiring structured variable abstraction. Task-specific fine-tuning altered the relative performance of model classes, with generative models outperforming encoders after adaptation. Our findings support the development of text-based prognostic tools that integrate seamlessly into clinical workflows without manual data extraction.

**Keywords:** Artificial Intelligence; Natural Language Processing; Clinical Decision Support; Electronic Health Records; modified Rankin Scale

**Non-standard Abbreviations and Acronyms:** *ASTRAL*, Acute Stroke Registry and Analysis of Lausanne; *BERT*, Bidirectional Encoder Representations from Transformers; *CI*, confidence interval; *CORAL*, COntinuous RAnked Logits; *CT*, computed tomography; *EHR*, electronic health record; *EMR*, electronic medical record; *GWTG*, Get With The Guidelines; *H&P*, history and physical; *HPI*, history of present illness; *IQR*, interquartile range; *IV*, intravenous; *LLM*, large language model; *LoRA*, Low-Rank Adaptation; *MAE*, mean absolute error; *MER*, mechanical thrombectomy; *MRI*, magnetic resonance imaging; *mRS*, modified Rankin Scale; *NIHSS*, National Institutes of Health Stroke Scale; *PHI*, Protected Health Information; *PLAN*, Prestroke modified Rankin Scale, Level of consciousness, Age, NIHSS; *AUPRC*, area under the precision–recall curve; *AUROC*, area under the receiver operating characteristic curve

**Introduction**

Stroke is a leading cause of disability worldwide, with the modified Rankin Scale (mRS) serving as the standard clinical measure for assessing post-stroke functional outcomes[1]. Accurate prediction of mRS scores can aid in clinical decision-making, goal setting, and resource allocation in stroke care. Traditionally, stroke outcome prediction has relied on structured clinical data such as National Institutes of Health Stroke Scale (NIHSS) scores[2–5], age[2–5], sex[5,6], comorbidities[5,7,8], laboratory values[5,9], length of stay[10], and discharge disposition[10–12]. These models can be effective, but they often require resource-intensive data abstraction and complex preprocessing, making them difficult to deploy in practice.[13,14] Additionally, the majority of these studies depend on information not available at admission, such as discharge NIHSS[4] and mRS[11,15] scores, disposition[10–12], or even later assessments such as 1-week[16] or 30-day mRS[17], in order to predict 90-day mRS scores. By contrast, models that use unstructured clinical notes from admission support low-friction deployment and provide prognostic insight early in the hospitalization to support clinical care.[14]

Unstructured clinical text has been shown to improve the prediction of functional outcomes after strokes. One study demonstrated that incorporating History of Present Illness (HPI) notes and CT reports significantly improved established baseline scores such as PLAN and ASTRAL.[18] Another study showed that using ClinicalBERT on the admission HPI performs similarly to using NIHSS for binarized 90-day mRS prediction.[19] A third study compared the predictive power of clinical text information, radiomics features, and DeepSurv-derived survival features using machine learning classifiers and found that text was the strongest single predictor of 90-day functional outcome.[20] Unstructured clinical text carries rich prognostic signal—often matching or exceeding the predictive value of traditional structured variables—while being readily available at admission and requiring minimal preprocessing or workflow disruption.

However, the optimal approach for extracting prognostic information from text remains unclear.[21,22] Large language models (LLMs) have recently shown strong promise for analyzing complex clinical text.[14,23] Unlike earlier natural language processing models that require extensive feature engineering, LLM encoders, such as BERT[24], can automatically generate contextualized representations that capture semantic relationships across text[24], whereas generative LLMs, such as GPT[25], can perform complex multi-step reasoning without task-specific training[25]. This makes them particularly well-suited for outcome prediction from heterogeneous and unstructured clinical text.[14,23] Yet, to date, no study has evaluated how generative LLMs perform on mRS prediction using unstructured clinical documentation as the sole input.

We conducted a comprehensive evaluation of LLM architectures, pretraining approaches, and fine-tuning configurations for predicting both the discharge and the 90-day mRS scores. An overview of our approach is displayed in **Figure 1**. Analyses were performed using clinical data from 10,302 stroke patient admissions collected at NYU Langone Health as part of the *Get With The Guidelines* initiative[26,27]. We assessed encoders (BERT[24] and NYUTron[14]) and generative models (Llama-3.1-8B[28] and MedGemma-4B[29]) under both frozen and fine-tuned conditions. These model choices allow us to directly compare generalist and clinically pretrained models of comparable size. We also reproduced the structured-data approach of Zhang et al.[4] and the combined structured- and unstructured-data approach of Sung et al.[19] to provide meaningful baseline comparisons against established 90-day mRS predictive models. This work seeks to clarify the potential and limitations of LLMs for predicting post-stroke functional outcomes and to inform their future integration into clinical decision-making.

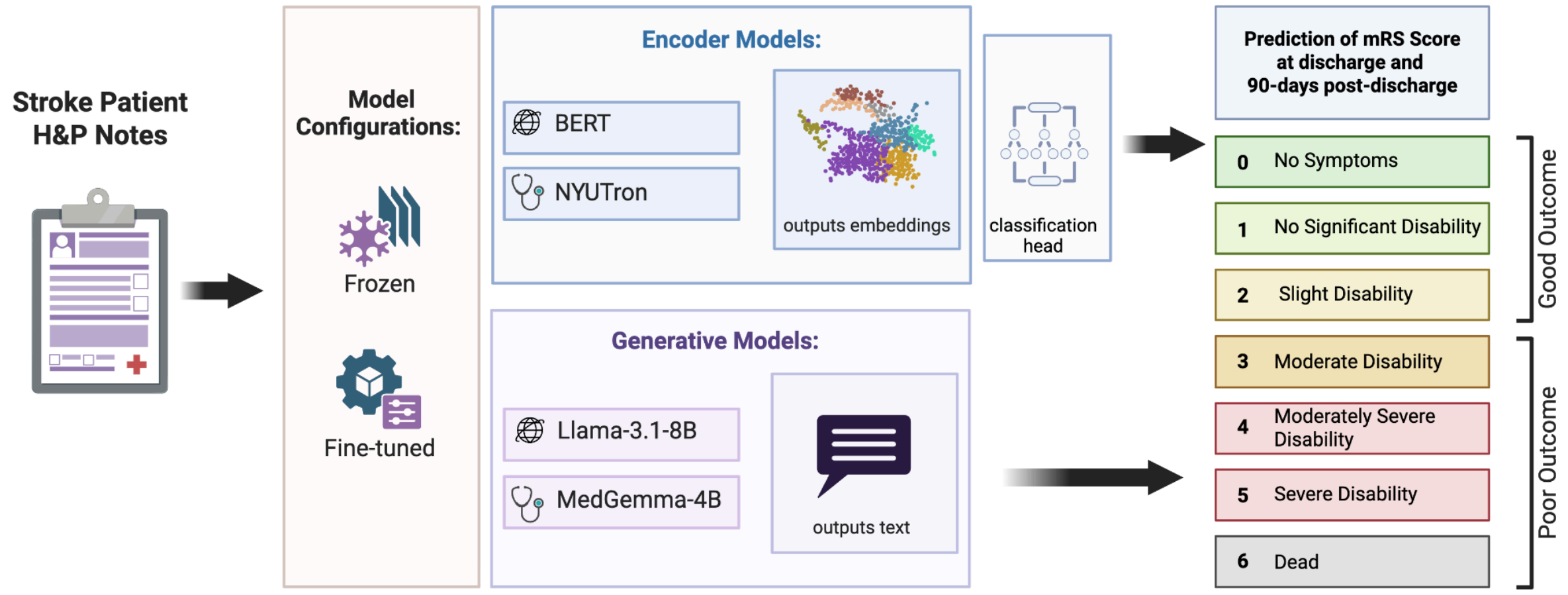

**Figure 1. Overview of stroke outcome prediction.** H&P notes from stroke patients were input into frozen and fine-tuned encoder and generative models to predict the mRS score at discharge and 90-day. The predicted mRS scores range from 0 (no symptoms) to 6 (death), with scores 0–2 binarized to represent good outcomes and 3–6 to represent poor outcomes.
Abbreviations: *BERT*, Bidirectional Encoder Representations from Transformers; *H&P,* History and Physical; *mRS*, modified Rankin Scale

## Methods

*Study Design*

This study was approved by the NYU Institutional Review Board (i22-0302). The need for informed consent was waived. The study was reported in accordance with the Transparent Reporting of Multivariable Prediction Model for Individual Prognosis or Diagnosis using Artificial Intelligence (TRIPOD+LLM) guidelines[30]. Data cannot be shared due to PHI, but code can be shared upon request. The corresponding author had full access to all study data and takes responsibility for the integrity of the data and the accuracy of the data analysis.

The study aims to predict both the discharge mRS and 90-day post-discharge mRS scores utilizing admission H&P notes, representing routine clinical documentation that captures pertinent patient information at the time of admission. All consecutive patients admitted for acute ischemic stroke between January 2016 and February 2025 in the NYU Langone Health (NYULH) Hospital System were identified from an institutional Get With The Guidelines–Stroke (GWTG-Stroke) registry[31]. NYULH is a university-affiliated medical center comprised of three certified Comprehensive Stroke Centers, located at NYU Langone Tisch Hospital in Manhattan, NYU Langone Hospital—Brooklyn, and NYU Langone Hospital—Long Island. All cases of stroke were prospectively registered in the stroke registries based on the following criteria: 1) patients with a diagnosis of stroke at the emergency department or during hospitalization, 2) daily screening of all patients receiving head CT, or 3) screening for a diagnosis at discharge using the International Classification of Diseases, Ninth and Tenth Revisions (ICD‐9 and ICD‐10 revisions with clinical modification) codes. Data regarding the demographics, cause, risk factor profiles, intervention, and outcomes of patients with stroke were collected. Stroke severity was assessed using the NIHSS and functional outcome was measured using the mRS.

Only cases with complete mRS labels (discharge and/or 90-day post-discharge) and valid H&P notes were included in this study. A flowchart with inclusion and exclusion criteria is provided in **Figure 2**. We extracted their discharge and 90-day post-discharge mRS scores from the registry and unstructured clinical notes (all H&P notes at admission) from the electronic medical record (EMR) via NYUMC Neuro Data Hub[32].

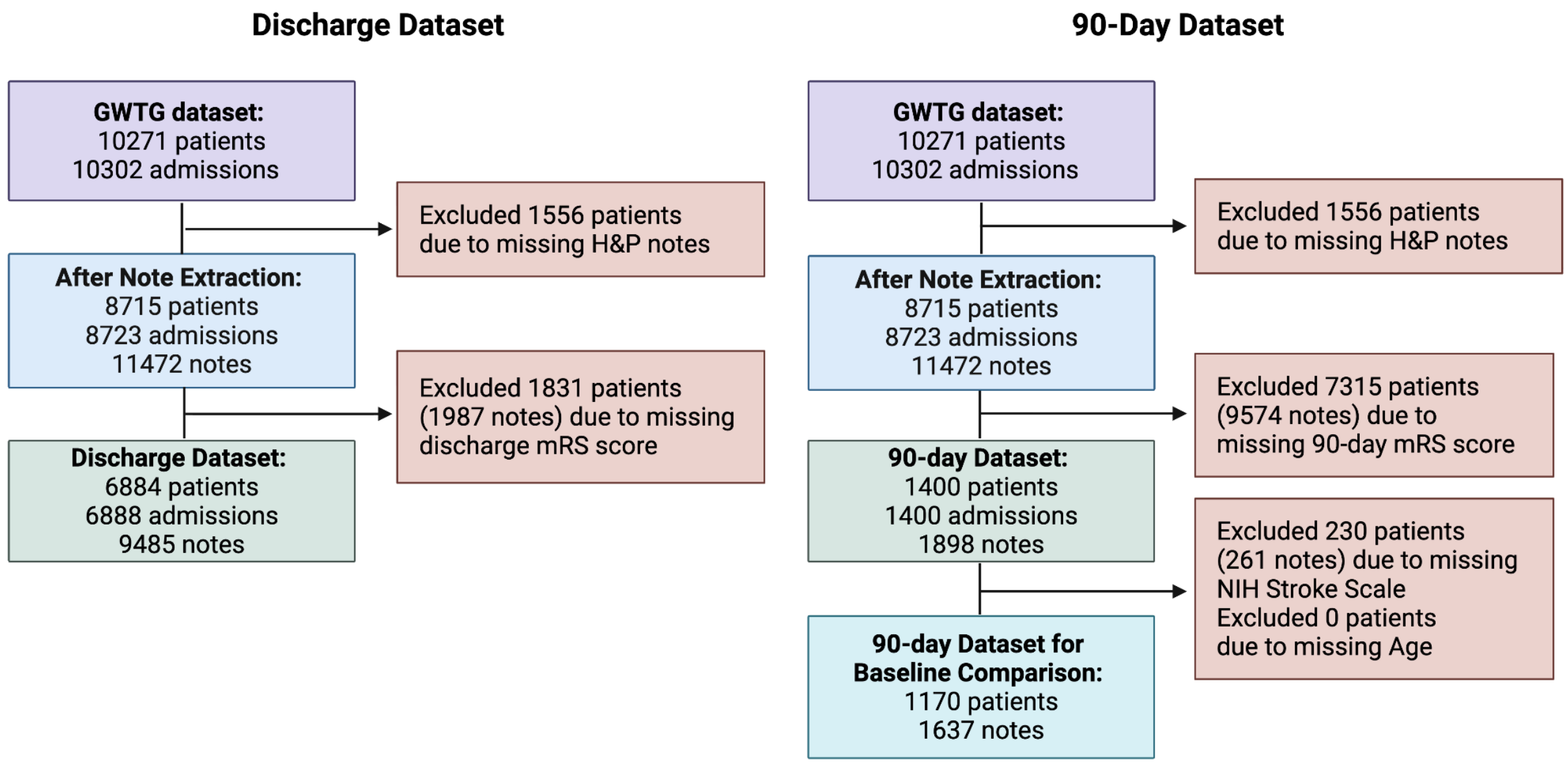

**Figure 2: Flowchart depicting construction of the discharge and 90-day datasets derived from the GWTG–Stroke registry and linked admission H&P notes.** A total of 10302 stroke admissions corresponding to 10,271 patients from the registry were identified. For the discharge analysis, admissions missing the discharge mRS score or H&P note were excluded, yielding a Discharge Dataset of 9,485 notes from 6,884 patients. For the 90-day analysis, admissions missing the 90-day mRS score or H&P note were excluded, resulting in a 90-day Dataset of 1,898 notes from 1,400 patients. A subset of the 90-day dataset excluding cases missing NIHSS values produced the 90-day Baseline Comparison Dataset (1,637 notes, 1,170 patients). These final cohorts were temporally split into train and test sets and were used for model development, cross-validation, and evaluation.
Abbreviations: *GWTG:* Get With The Guidelines*; H&P,* History and Physical; *mRS*, modified Rankin Scale; *NIHSS*, National Institutes of Health Stroke Scale

*Outcomes*
The primary outcomes were the exact 90-day mRS score (total 7 classes: 0, 1, 2, 3, 4, 5, and 6) and good vs. poor functional outcome after acute stroke (mRS 0-2 vs. 3-6). The secondary outcomes were the exact discharge mRS score (total 7 classes: 0, 1, 2, 3, 4, 5, and 6) and good vs. poor functional status at discharge (mRS 0-2 vs. 3-6).

*Data*
Patients with available H&P notes and corresponding mRS scores were included in the discharge and/or 90-day datasets, depending on which outcome timepoints were available. The discharge dataset consists of 9,485 H&P notes corresponding to 6,888 admissions and 6,884 patients. The 90-day post-discharge dataset, which is restricted to patients who underwent thrombectomy or thrombolysis, consists of 1,898 notes corresponding to 1,400 admissions and 1,400 patients. For patients with multiple admission notes, each note was treated as an independent sample (**Supplementary Fig. S1**). To simulate prospective real-world deployment, we adopted a temporal split in which the most recent 12 months of admissions were reserved as the held-out test set. This yielded 8,408 training/validation notes and 1,077 test notes for the discharge dataset, and 1,674 training/validation notes and 224 test notes for the 90-days post-discharge dataset (train/test split: 90/10).

*Preprocessing*
H&P notes were preprocessed using each model's native tokenizer, which segments text into model-specific tokens (e.g., words or subwords). To mimic real-world deployment scenarios, no manual text selection or processing was performed. **Figure S2** displays the distribution of H&P token lengths across all admission notes.

For encoder models (BERT[24] and NYUTron[14]), we truncated the H&P notes to a maximum of 512 tokens (approximately two paragraphs, roughly corresponding to the HPI) to conform to the input length limitations of these models; after testing three different strategies of converting the H&P notes into 512 tokens—first 512, last 512, and summarization— we chose to truncate each note to the first 512 tokens, as depicted in **Figure S3**. Meanwhile, generative models (Llama[28] and MedGemma[29]) had larger context windows, enabling input sequences up to 8192 tokens.

*Clinical Features*
To compare with other baseline prognostic models, we additionally collected age and NIHSS score from the structured registry dataset for 90-day mRS prediction and reproduced the Sung et al. model[19] and the Zhang et al. model[4]**.**

*Encoder Models*
We evaluated two encoder models: BERT[24] and NYUTron[14]. BERT[24] is a widely used general-purpose encoder transformer model pretrained on large corpora of text. NYUTron[14] is a clinical adaptation of the BERT architecture that is specifically pretrained on a corpus of hospital notes.

Both models were evaluated in frozen and fine-tuned configurations. In the frozen configuration, we used the embeddings produced by the models as input for downstream prediction tasks. We compared four classifier heads: linear regression, L2-regularized logistic regression, CORAL ordinal regression, and a random forest classifier. For each encoder, the classifier head that achieved the highest performance during cross-validation was selected and subsequently applied to the held-out test set, as depicted in **Figure S4**. In the fine-tuned configuration, we fine-tuned BERT and NYUTron for the specific task of mRS prediction. We systematically unfroze varying numbers of transformer layers—from 0 (frozen embeddings) to 12 (full fine-tuning)—and identified the optimal fine-tuning depth for each mode (**Fig. S5**). The embeddings were passed through a linear regression head to predict continuous mRS scores (range 0-6). We optimized the model using mean absolute error (MAE) loss. Predictions were clipped to the 0-6 range to match the mRS scale and then rounded to the nearest integer for discrete mRS classification. All fine-tuned encoder models were trained on a single NVIDIA A100 GPU with a consistent optimization setup of five epochs, batch size = 8, optimizer = AdamW, learning rate = $2\times10^{-5}$.

*Generative Models*

We also evaluated two generative LLMs: Llama-3.1-8B[28] and MedGemma-4B[29]. As autoregressive transformers, these models generate text sequentially while drawing on information from across the entire H&P note.[25] Llama-3.1-8B is a general-purpose model pretrained on vast internet data[28], whereas MedGemma-4B is a clinical model pretrained on medical text[29]. Both models were evaluated using zero-shot prompting and parameter-efficient fine-tuning.

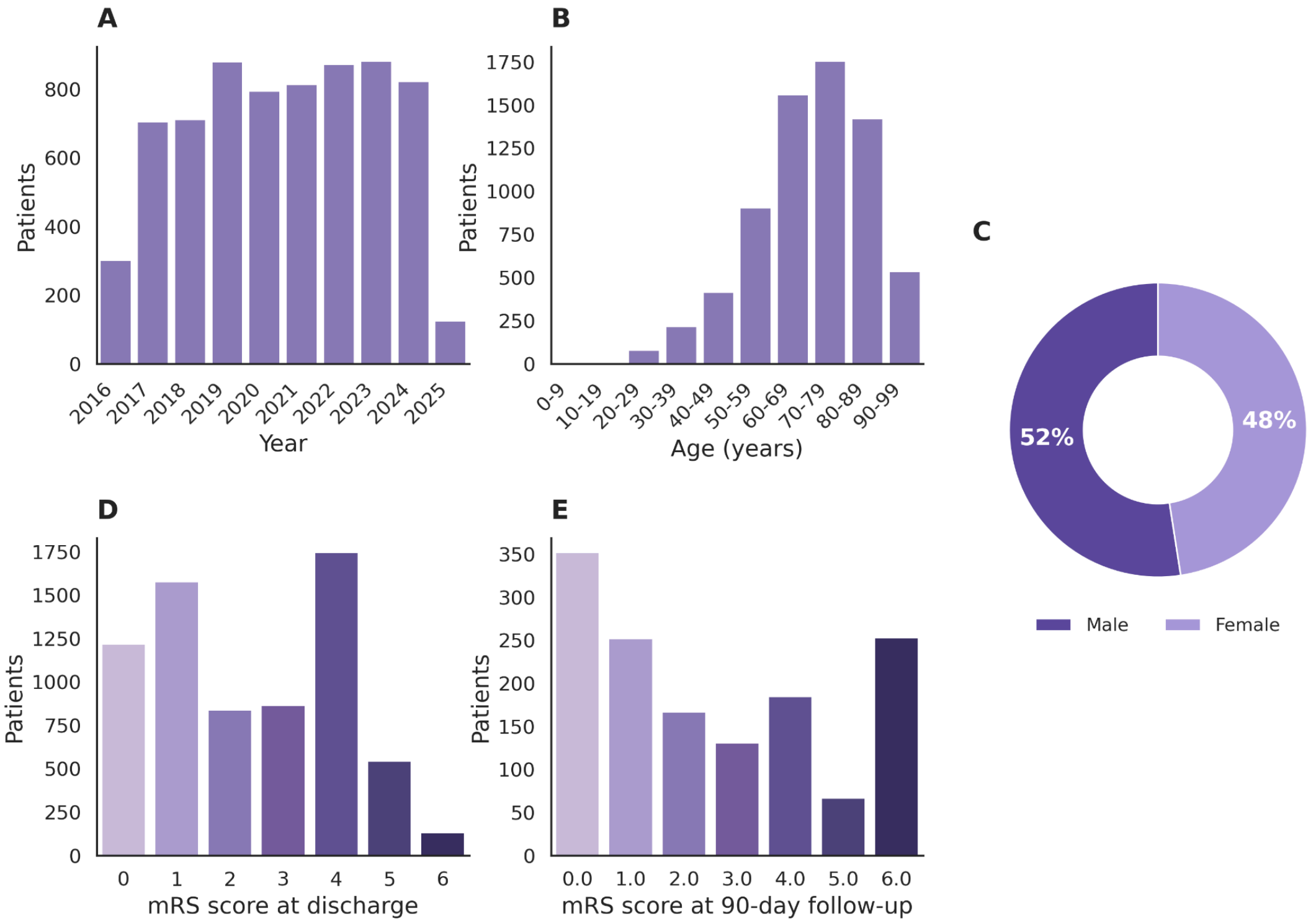

**Figure 3. Overview of the stroke cohort (Discharge: N = 6,888 admissions; 90-day: N = 1,400 admissions).** Panels A–D summarize characteristics of the discharge cohort at the admission level—including admission counts by year (A), age distribution (B), sex distribution (C), and discharge mRS (D). Panel E depicts the 90-day mRS distribution for admissions with available follow-up (limited to those treated with thrombectomy or thrombolysis).

For the zero-shot prompting, we used the -Instruct variants[28,29] of the models. We tested two versions of the prompt, provided in **Figure S6**: Prompt A only instructed the model to predict the mRS score, whereas Prompt B also provided the mRS definitions. We found no significant difference between two prompts in our experiments, suggesting that the models are familiar with the concept of mRS (**Fig. S7**). Results throughout the study are reported for Prompt B.

For fine-tuning, we used the base variants of each model. We converted them into seven-class mRS classifiers by attaching a task-specific softmax classification head. We used Low-Rank Adaptation (LoRA)[33] to efficiently fine-tune the models with a reduced number of trainable parameters. All generative models were fine-tuned on a single NVIDIA A100 GPU using an identical set of hyperparameters: three epochs, batch size of 1 with four gradient-accumulation steps for an effective batch size of 4, AdamW optimizer, with a learning rate of $5\times10^{-5}$ and weight decay of 0.01. We used a uniform LoRA configuration ($r = 16$, $\alpha = 32$, dropout = 0.05, target modules = [q_proj, v_proj]).

*Baseline Prognostic Models*

We reproduced two established 90-day mRS prediction baseline models: the structured-data model from Zhang et al.[4] and the text-augmented structured model from Sung et al.[19]

For the structured-data baseline, we implemented the ordinal logistic regression model introduced by Zhang et al.[4]. Their model identified age and NIHSS as independent predictors of 90-day mRS. Following their approach, we trained a proportional-odds regression model using these two variables.

For the hybrid baseline, we replicated the best-performing architecture described by Sung et al.[19], which integrates unstructured clinical text with key structured predictors. In their approach, models using the HPI text were enhanced by combining text-derived predictions with age and NIHSS. Without having their model weights, we approximated this setup by fine-tuning Bio-ClinicalBERT on H&P notes to generate predicted probabilities of poor outcome (mRS $\geq$ 3) and then incorporating these probabilities together with age and NIHSS into a logistic regression classifier.

Our best-performing model—fine-tuned Llama—was compared against these two baseline models. For exact 90-day mRS accuracy, only the models producing ordinal or multiclass mRS outputs (Zhang et al. and Llama) were evaluated, as Sung et al. report binary outcomes only. All three models (Zhang et al., Sung et al., and Llama) were also assessed on binary accuracy (mRS 0–2 vs. 3–6).

*Evaluation Metrics*

Model performance was assessed using both exact and binary accuracy. Exact accuracy was defined as the proportion of predictions in which the model's mRS output (total 7 classes: 0-6) matched the true discharge or 90-day mRS score exactly. Binary accuracy was defined as the proportion of predictions in which the model's mRS output (total 7 classes: 0-6) matched the favorable outcome category (mRS scores ranging 0–2) or the unfavorable outcome category (mRS scores ranging 3–6).

**Table 1. Baseline demographic and clinical characteristics of the discharge and 90-day cohorts.**

| Characteristic | Discharge cohort (N = 6,888 admissions) | 90-day cohort (N = 1,400 admissions) |
|---|---|---|
| Age (median [IQR]) | 71 [61-81] | 73 [61-83] |
| Sex | Female, 47.6%<br>Male, 52.4% | Female, 48.5%<br>Male, 51.5% |
| Admission NIHSS (median [IQR]) | 3 [1-8]<br>(Excluding missing data for 1.1% of patients) | 10 [4-19]<br>(Excluding missing data for 3.3% of patients) |
| Discharge Disposition | Home, 55.0%<br>Other Health Care Facility, 38.5%<br>Acute Care Facility, 0.6%<br>Left AMA, 0.8%<br>Hospice - Health Care Facility, 2.9%<br>Hospice - Home, 0.7%<br>Expired, 1.6% | Home, 44.8%<br>Other Health Care Facility, 41.0%<br>Acute Care Facility, 0.5%<br>Left AMA, 0.4%<br>Hospice - Health Care Facility, 7.7%<br>Hospice - Home, 0.7%<br>Expired, 3.9%<br>(Excluding missing data for 1% of patients) |
| IV Thrombolytic Initiated | 11.5%<br>(Excluding missing data for .4% of patients) | 52.0%<br>(Excluding missing data for 1.5% of patients) |
| Mechanical Thrombectomy (MER) Initiated | 14.2%<br>(Excluding missing data for 12.5% of patients) | 58.1%<br>(Excluding missing data for 3.6% of patients) |
| Final Stroke Diagnosis | Ischemic stroke only | Ischemic stroke only |
| mRS (median [IQR]) | Discharge mRS: 2 [1-4] | 90-day mRS: 2 [0-4] |
| Binary mRS | Discharge mRS $< 3$, 52.6%<br>Discharge mRS $\geq 3$, 47.4% | 90-day mRS $< 3$, 54.9%<br>90-day mRS $\geq 3$, 45.1% |

Abbreviations: *AMA*, against medical advice; *IQR*, interquartile range; *IV,* intravenous; *MER*, mechanical thrombectomy; *mRS*, modified Rankin Scale; *NIHSS*, National Institutes of Health Stroke Scale

*Statistical Analysis*
Metrics were reported as point estimates, with cross-validation results averaged across folds and test-set performance bootstrapped (1000 resamples) to obtain 95% confidence intervals. Model comparisons were performed on paired predictions using McNemar's test. All tests were two-sided, with the significance level of 0.05. Statistical analyses were conducted in Python (v3.9) using SciPy, scikit-learn, statsmodels, and PyTorch (2.7.1) with Hugging Face Transformers.

## Results

*Study population*
A total of 10,271 patients were registered to the study cohort during the study period. After excluding 1556 patients with missing H&P notes, 8715 patients were included. Of these, 1831 patients (21%) were missing discharge mRS scores and 7315 patients (84%) were missing 90-day post-discharge mRS scores. Patients with available mRS scores were allocated to two groups: discharge mRS (N=6884) and 90-days post-discharge mRS (N=1400). Baseline demographics and clinical characteristics are detailed in **Table 1** and **Figure 3**. Among the 1400 patients in the 90-day dataset, approximately 52.0% received IV thrombolytic therapy, and 58.1% received thrombectomy (**Table 1**). The distribution of mRS scores at discharge and at 90 days post-discharge is provided in **Figure 3D** and **3E**.

*Discharge mRS prediction with frozen models*
As displayed in **Figure 4**, in the frozen configuration, the clinically pretrained NYUTron performed best. For the exact discharge mRS prediction, NYUTron achieved the highest accuracy of 35.6% [95% CI, 32.8%-38.4%], significantly outperforming general-purpose BERT at 30.6% [95% CI, 28.0%-33.2%] ($P = 2.32\times10^{-5}$), MedGemma at 25.6% [95% CI, 23.0%-28.2%] ($P = 5.62x10^{-8}$), and Llama at 17.6% [95% CI, 15.3%-19.9%] ($P = 5.31x10^{-17}$). Both MedGemma and Llama were also significantly outperformed by BERT ($P = 7.43x10^{-3}$ and $P = 1.88x10^{-10}$, respectively).

For the binary discharge mRS prediction, NYUTron (72.0% [95% CI, 69.3%-74.7%]), significantly outperformed MedGemma (66.4% [95% CI, 63.5%-69.3%], $P = 2.61x10^{-3}$), Llama (65.2% [95% CI, 62.4%-68.0%], $P = 2.61x10^{-3}$), and BERT (59.8% [95% CI, 56.9%-62.7%]; $P = 1.17\times10^{-13}$). Consistent with these results, NYUTron achieved the highest area under the receiver operating characteristic curve (AUROC) of 0.79 and area under the precision–recall curve (AUPRC) of 0.78 (**Fig. S8A and S8B**).

*90-day mRS prediction with frozen models*
90-day mRS prediction results followed a similar pattern. NYUTron again achieved the highest exact accuracy of 31.7% [95% CI, 25.7%-37.7%], followed by BERT (28.6% [95% CI, 22.8%-34.4%]), MedGemma (24.1% [95% CI, 18.7%-29.5%]), and Llama (20.1% [95% CI, 15.0%-25.2%]). Only the NYUTron versus Llama comparison reached statistical significance ($P = 1.15\times10^{-2}$).

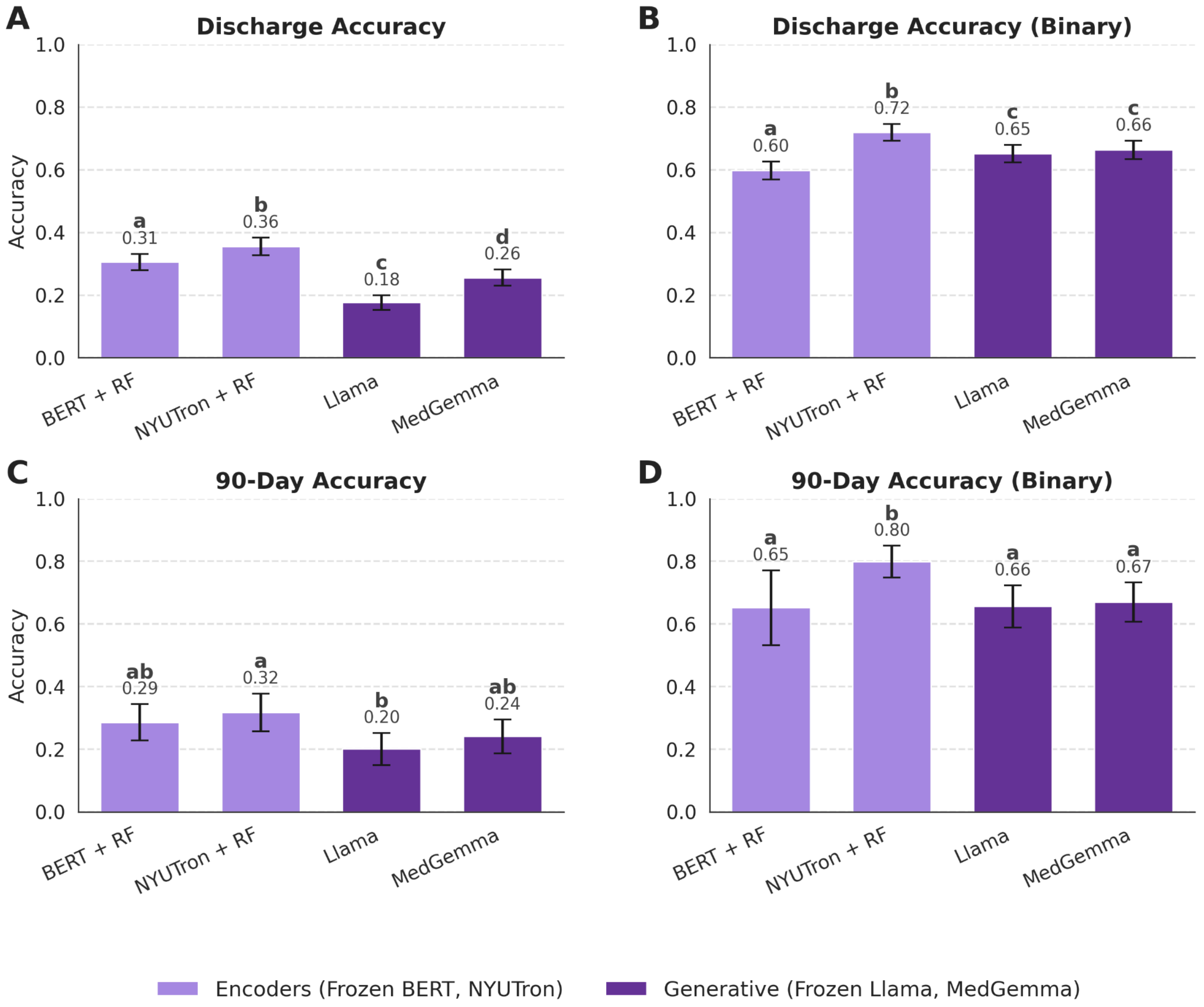

**Figure 4. Frozen model comparison for discharge and 90-day mRS prediction.** Panels A–D show exact accuracy (left column) and binary accuracy (right column; mRS 0–2 vs 3–6) with 95% confidence intervals derived from bootstrapped test-set estimates for four frozen models evaluated on discharge (Panels A–B) and 90-day (Panels C–D) outcomes. Encoder models (BERT and NYUTron) use random-forest classifier heads trained on frozen embeddings, whereas generative models (Llama and MedGemma) are evaluated in their frozen zero-shot configurations using the definition-augmented prompt. Shared significance-group lettering (a–d) above each bar denotes CLD groupings: models that do not share a letter differ significantly ($p < 0.05$), whereas models that share at least one letter are not significantly different.

Abbreviations: *CLD*, compact letter display; *mRS*, modified Rankin Scale; *RF*, random forest

For binary 90-day accuracy, the pattern was similar to that of the binary discharge mRS prediction. NYUTron achieved the highest accuracy of 79.9% [95% CI, 74.8%-85.0%], significantly outperforming MedGemma (67.0% [95% CI, 60.7%-73.3%], $P = 1.05\times10^{-3}$), Llama (65.6% [95% CI, 58.9%-72.3%], $P = 1.31\times10^{-4}$), and BERT (65.2% [95% CI, 53.3%-77.1%]; $P = 2.25\times10^{-5}$). No other comparisons were found to be significant. NYUTron again exhibited superior AUROC of 0.86 and AUPRC of 0.80 (**Fig. S8C and S8D**).

*Discharge mRS prediction with fine-tuned models*

As shown in **Figure 5**, after fine-tuning, generative models outperformed encoders for the discharge mRS prediction. For the exact mRS prediction, MedGemma achieved the highest accuracy of 42.7% [95% CI, 39.7%-45.7%], followed by Llama at 42.0% [95% CI, 39.0%-45.0%], NYUTron at 32.9% [95% CI, 30.0%-35.8%], and BERT at 27.4% [95% CI, 24.8%-30.0%]. All pairwise comparisons between generative and encoder models reached statistical significance (all pairwise $P < 9.08\times10^{-8}$).

For the binary discharge functional outcome accuracy, Llama performed best at 75.0% [95% CI, 72.4%-77.6%], followed by MedGemma (74.7% [95% CI, 72.1%-77.3%]), NYUTron (74.6% [95% CI, 72.0%-77.2%]), and BERT (68.9% [95% CI, 66.1%-71.7%]). BERT performed worse than all other models (all pairwise $P < 1.07\text{x}10^{-4}$), but no other comparisons were significant. In line with these findings, Llama and MedGemma achieved higher AUROC and AUPRC (Llama: AUROC = 0.84; AUPRC = 0.81; MedGemma: AUROC = 0.84; AUPRC = 0.82) than NYUTron and BERT (NYUTron: AUROC = 0.81; AUPRC = 0.79; BERT: AUROC = 0.74; AUPRC = 0.72) (**Fig. S9A and S9B**).

*90-day mRS prediction with fine-tuned models*

For the exact 90-day mRS prediction, Llama achieved the highest accuracy of 33.9% [95% CI, 27.9%-39.9%], followed by MedGemma at 31.7% [95% CI, 25.5%-37.9%], NYUTron at 25.4% [95% CI, 19.8%-31.0%], and BERT at 25.0% [95% CI, 19.2%-30.8%]. Only Llama significantly outperforms encoders ($P = 3.82\text{x}10^{-2}$ and $P = 4.28\text{x}10^{-2}$ against BERT and NYUTron, respectively).

For the binary 90-day functional outcome prediction, performance differences were modest and no pairwise comparisons between generative and encoder models reached statistical significance. NYUTron was highest at 76.8% [95% CI, 71.8%-81.8%], followed by Llama (76.3% [95% CI, 70.7%-81.9%]), MedGemma (75.9% [95% CI, 70.5%-81.3%]), and BERT (70.1% [95% CI, 64.1%-76.1%]). However, Llama achieved the highest AUROC and AUPRC (AUROC = 0.87; AUPRC = 0.84) (**Fig. S9C and S9D**).

*Error analysis*

To characterize model behavior across the full mRS spectrum, we examined confusion matrices for each model. The confusion matrix for the best-performing fine-tuned model, Llama, is presented in **Figure S10**. These matrices exhibit a clear diagonal pattern, with 55.0% of discharge and 39.19% of 90-day misclassifications occurring within ±1 mRS level, demonstrating a tendency toward near-neighbor rather than large-magnitude errors.

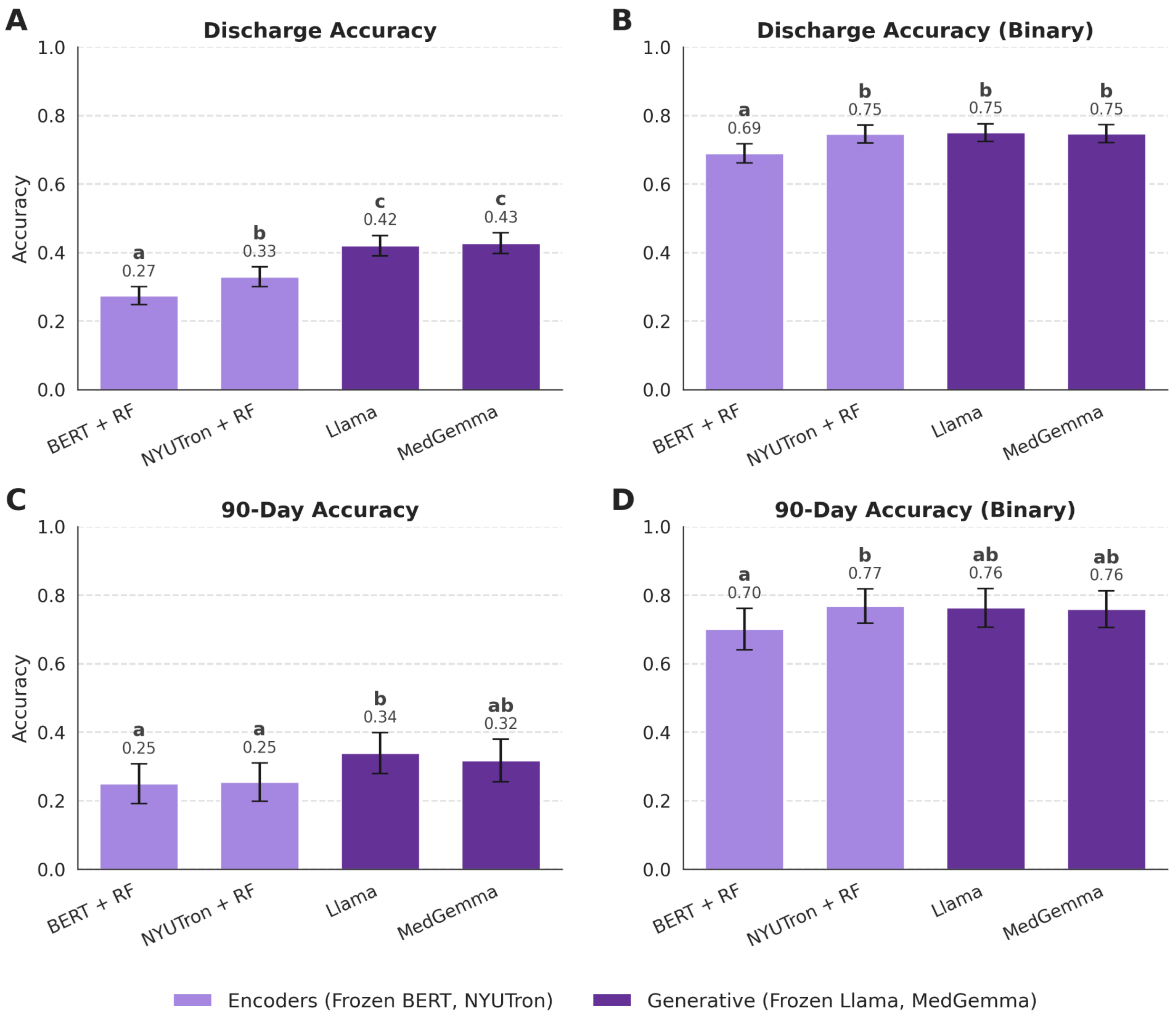

**Figure 5. Fine-tuned model comparison for discharge and 90-day mRS prediction.** Panels A–D show exact accuracy (left column) and binary accuracy (right column; mRS 0–2 vs 3–6) with 95% confidence intervals derived from bootstrapped test-set estimates for four fine-tuned models evaluated on discharge (Panels A–B) and 90-day (Panels C–D) outcomes. All models were fine-tuned on patient admission H&P notes. Encoder models (BERT and NYUTron) were fine-tuned to a depth of 10 layers—identified as the most effective depth across both encoders and timepoints—using linear classification heads trained jointly with the fine-tuned encoder. Generative models (Llama and MedGemma-4B) were fine-tuned using LoRA. Shared significance-group lettering (a–d) above each bar denotes CLD groupings: models that do not share a letter differ significantly ($p < 0.05$), whereas models sharing at least one letter are not significantly different.

Abbreviations: *CLD*, compact letter display; *H&P,* History and Physical; *LoRA*, Low-Rank Adaptation; *mRS*, modified Rankin Scale, *RF*, random forest

*Structured-data baselines comparison*

The best-performing model—fine-tuned Llama— achieved performance comparable to established baselines, including the NIHSS+age ordinal-regression model of Zhang et al.[4] and the HPI+NIHSS+age binary-classification model of Sung et al[19] (**Fig. 6**). For the exact 90-day mRS accuracy, Llama achieved 34.6% [95% CI, 27.9%-41.3%] compared with 28.1% [95% CI, 21.8%-34%, P = .08] for the proportional-odds baseline of Zhang et al.[4] For binary 90-day mRS accuracy, Llama again achieved the highest performance (77.1% [95% CI, 70.9%-83.2%]), followed by the Sung et al. (73.8% [95% CI, 67%-80.4%], P = .85) and the Zhang et al. models[4] (72.6% [95% CI, 65.9%-78.8%], P = .43).

## Discussion

We present the first comprehensive evaluation of modern LLMs for fine-grained mRS prediction using unstructured clinical text from early admission. While the majority of past studies were limited to datasets of fewer than 500 patients[4,11,34], we employed a dataset of over 6,000 stroke admissions from the Get With The Guidelines–Stroke registry initiative. Unlike prior approaches that relied on structured variables[5], manual text feature selection[35], or post-discharge data[4,11], our method leverages unmodified H&P notes—routinely created at every admission—to predict both discharge and 90-day functional outcomes. This task has some similarities to but is substantially different from mRS identification, where the current score is inferred from the note, rather than predicted into the future.[35] We demonstrated that text-based models can achieve predictive performance comparable to established structured-data models, offering a practical and scalable foundation for clinical decision support.

Unlike most prior work, which reports either binary[10,11,18,19,36] or one-off accuracy[4,37], we evaluated both exact and binarized mRS prediction accuracies to better characterize model behavior. In doing so, we observed that relative performance differences between models were more pronounced for exact mRS accuracy than for binary accuracy, reflecting the greater discriminative resolution of ordinal prediction compared with coarse dichotomization. Additionally, fine-grained 0-6 mRS predictions provide actionable patient-specific prognostic detail for counseling and rehabilitation planning.[4,37] Although exact accuracy was modest—as expected for a seven-class ordinal outcome—error analysis revealed that most misclassifications fell within ±1 mRS point, indicating near-neighbor rather than clinically implausible errors.

Our findings extend and clarify mixed results from prior text-based stroke outcomes prognostication studies. Heo et al. (2019) concluded that deep learning outperformed traditional machine learning for mRS prediction from MRI reports[21], whereas Su et al. (2025) found the opposite using admission notes[22]. Our fine-tuned Llama model achieved numerical improvements over both the proportional-odds model inspired by Zhang et al. (2021)[4] and the structured-plus-text approach similar to that of Sung et al. (2021)[19]. While these differences did not reach statistical significance in our 90-day cohort, the results suggest that early clinical narratives contain sufficient prognostic signals to inform long-term functional outcome prediction. Text-based prediction avoids the resource-intensive data abstraction that limits structured-variable approaches[13,14], offering a lower-friction path to deployment.

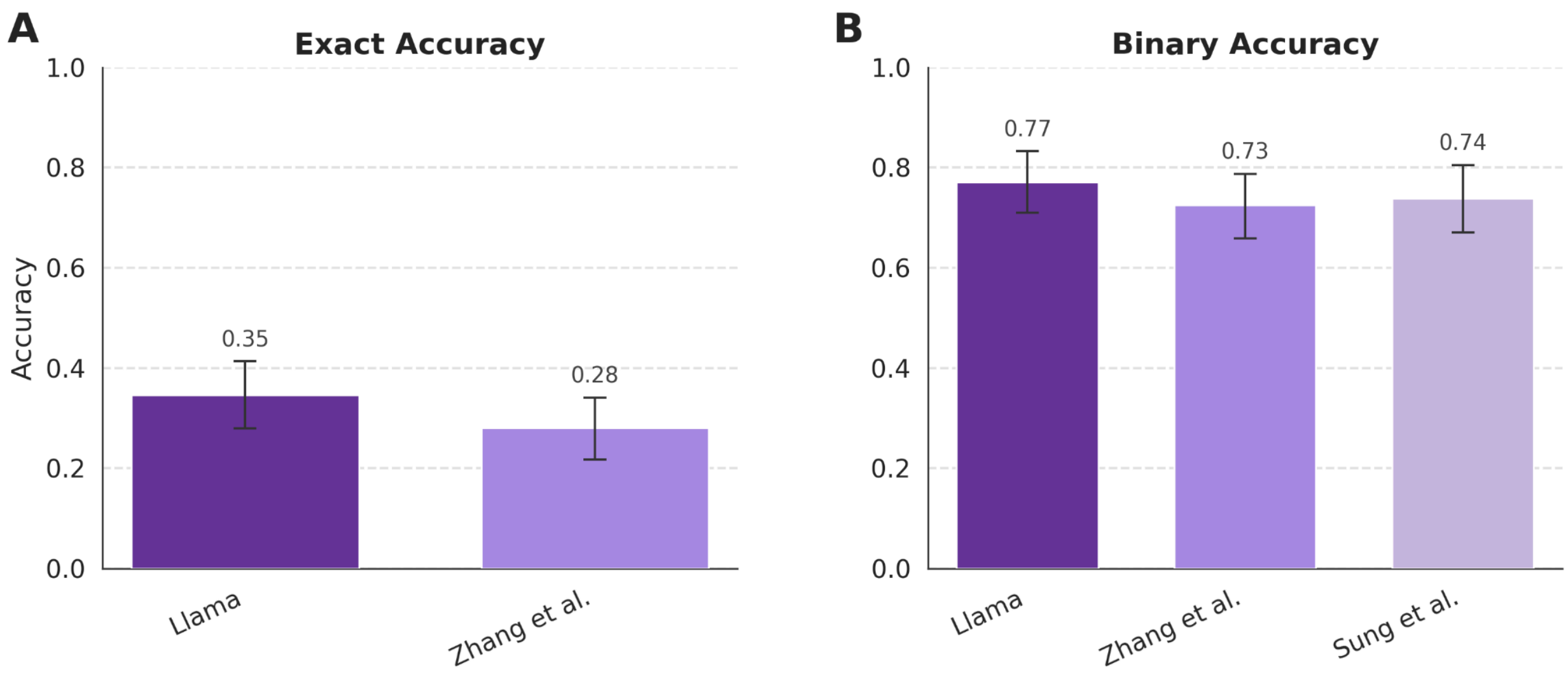

**Figure 6: Comparison of fine-tuned Llama-3.1-8B and baseline models for 90-day mRS prediction.** Panels A–B present mean predictive performance for three prior approaches: a fine-tuned Llama model, the proportional-odds ordinal regression model from Zhang et al., and the structured-plus-text binary classifier from Sung et al. (A) Exact accuracy for 90-day mRS prediction comparing Llama with the ordinal regression model; the Sung et al. model is omitted from this panel because it produces binary-only outputs. (B) Binary accuracy for predicting poor functional outcome (mRS $\geq$ 3) across all three models. Error bars denote bootstrapped 95% confidence intervals computed on the 90-day test set restricted to patients with available NIHSS and age.

Abbreviations: *mRS*, modified Rankin Scale, *NIHSS*, National Institutes of Health Stroke Scale

It is important to note that our implementations of the Sung et al. and Zhang et al. baselines are similar but not exact replications of the original methods. Sung et al. used the HPI section of the admission note, whereas we approximated this by extracting the first 512 tokens of the H&P note. We also did not apply pre-processing steps, such as spell-correction and acronym expansion nor a hyperparameter search. Zhang et al. utilized discharge NIHSS while our implementation used the admission value. Unlike their study, we also did not need to upsample our dataset due to sufficient size. There is also a natural dataset shift compared to these methods: Sung et al.'s data originated from Taiwanese hospitals, which may have different documentation styles. Zhang et al.'s dataset also had specific inclusion criteria such as eligibility for endovascular therapy and a baseline NIHSS above 5, and excluded patients who died during hospitalization. Meanwhile, our 90-day dataset only consists of patients who underwent thrombectomy or thrombolysis. While the lack of exact weights and differences in methodology and datasets prevent a direct comparison to these studies, our findings suggest that fine-tuned Llama performs at least on par with structured and semi-structured approaches.

The relative performance of encoder versus generative architectures differed between frozen and fine-tuned conditions. Before fine-tuning, encoders tended to outperform generative models. However, this comparison is complicated by structural asymmetries: frozen encoders use trainable classifier heads that effectively function as a small task-supervised fine-tuning, whereas generative models rely on prompt-engineering alone. After fine-tuning, generative models achieved the highest exact mRS accuracies, which could reflect their capacity to model long-range dependencies across full H&P notes. Importantly, our encoders were limited to 512 tokens, whereas generative models processed sequences up to 8,192. We also did not control for the model parameter size between the encoder and generative models. Future work using long-context encoders such as ModernBERT[38] or BigBird[39] could clarify whether this performance gap reflects architectural differences, model weight class, or input length constraints.

Our results underscore the importance of task-specific pretraining and fine-tuning for clinical prediction tasks. In the frozen setting, NYUTron—the only model pretrained on clinical notes—achieved the highest accuracy, outperforming generative models that are an order of magnitude larger. This is consistent with recent findings showing that general-purpose foundation models underperform on EHR-based hospital operations tasks, such as readmission and length of stay predictions, without supervised adaptation.[23] Notably, for the binary discharge task, the generative models outperformed BERT, suggesting that NYUTron's advantage stems from its clinical pretraining rather than its encoder architecture. Furthermore, NYUTron remained consistently more accurate than BERT even after fine-tuning, matching generative models on the binary task, suggesting that domain-specific pretraining on clinical text confers a durable performance advantage that persists even after supervised fine-tuning.

However, we found that parameter-efficient fine-tuning via LoRA can effectively adapt general models to specialized clinical tasks. Once LoRA-fine-tuned, generative models achieved comparable binary accuracy and surpassed NYUTron in exact mRS prediction. This finding suggests that supervised adaptation can enable general-purpose models to match or exceed the performance of encoders pretrained on institutional EHR data, offering a practical alternative to developing and maintaining domain-specific pretrained models that require larger-scale datasets and substantial pretraining resources. In this study, we limited our investigation to smaller local models. As larger HIPAA-compliant generalist foundation models such as GPT-5 become increasingly available across health systems, comparing generalist and clinical specialist systems across tasks remains an important direction for future investigation.[23,40,41]

Our study is not without limitations. As mentioned above, we did not control for the model size when comparing encoder and generative models. We also did not incorporate imaging reports[18], which have been shown to contribute complementary prognostic information such as infarct volume and location. The choice to use a predetermined temporal split resulted in a limited size of the 90-day cohort with complete structured variables, potentially underpowering our comparison against established baselines. Additionally, as the 90-day cohort was limited to patients who underwent thrombectomy or thrombolysis and had documented follow-up, it's unclear whether our 90-day performance estimates will generalize to the broader stroke population. Lastly, although mRS is a well-established measure of functional outcome, it remains a subjective and coarse-grained scale that can introduce labeling variability.[42]

Several next steps emerge from our findings and limitations. Future work should pursue external validation across diverse hospital systems, explicitly evaluate equity and documentation-related biases, and assess robustness to variation in writing style and workflow. Multimodal approaches integrating text with imaging[41,43–45] or structured variables[5] may improve performance, though doing so necessitates careful consideration of admission-time data availability. Comparing trainable local models against HIPAA-compliant frontier models is important to establish the benefits of the biomedical and/or clinical pretraining and fine-tuning for this task.[23,40,41,46] Prospective studies assessing the impact on clinical decision-making and algorithmic fairness evaluation are essential before deployment.[47] Additionally, the development of lightweight AI systems demonstrates the potential for deploying efficient clinical prediction models directly on edge devices, which could further reduce infrastructure barriers and enable real-time prognostication at the point of care. Taken together, these directions can help translate text-based prognostic models from research settings into real-world clinical environments.

In conclusion, we evaluated modern large language models on a real-world clinical prediction task—forecasting post-stroke functional outcomes from admission documentation. Unlike standardized multiple-choice benchmarks commonly used to assess medical AI[47–49], this task reflects the complexity inherent in clinical prognostication: incomplete information, variable documentation quality, and outcomes shaped by factors not yet observable at admission. While EHR-pretrained NYUTron performed best in the frozen setting, task-specific fine-tuning enabled generative models to surpass encoders. Our findings demonstrate that meaningful stroke outcome prediction can be achieved using routinely documented clinical text alone, supporting the development of prognostic tools that integrate seamlessly into existing clinical workflows without requiring manual data extraction.

**Acknowledgements**: We would like to acknowledge Nader Mherabi, Dafna Bar-Sagi, PhD, and Paul Testa, MD for their continued support of medical AI research at NYU. We thank Michael Constantino, Kevin Yie, and the NYU Langone High-Performance Computing (HPC) Team for supporting computing resources fundamental to our work.

**Sources of Funding**: This work was supported by the Institute for Information & Communications Technology Promotion (IITP) grant funded by the Korea government (MSIT) (No. RS-2019-II190075 Artificial Intelligence Graduate School Program (KAIST); No. RS-2024-00509279, Global AI Frontier Lab).

**Disclosures:** EKO has equity in Delvi, MarchAI, and Artisight, income from Merck & Co. and Mirati Therapeutics, employment at Eikon Therapeutics, and consults for Sofinnova Partners and Google.

**Conflict of Interest**: None
**Supplemental Material:** Figures S1-S10

**References**

1. Saver, J. L. *et al.* Standardized nomenclature for modified Rankin Scale global disability outcomes: Consensus recommendations from Stroke Therapy Academic Industry Roundtable XI. *Stroke* **52**, 3054–3062 (2021).

2. Rost, N. S. *et al.* Stroke severity is a crucial predictor of outcome: An international prospective validation study. *J. Am. Heart Assoc.* **5**, (2016).

3. Wouters, A., Nysten, C., Thijs, V. & Lemmens, R. Prediction of outcome in patients with acute ischemic stroke based on initial severity and improvement in the first 24 h. *Front. Neurol.* **9**, 308 (2018).

4. Zhang, M. Y., Mlynash, M., Sainani, K. L., Albers, G. W. & Lansberg, M. G. Ordinal prediction model of 90-day modified Rankin Scale in ischemic stroke. *Front. Neurol.* **12**, 727171 (2021).

5. Klug, J., Leclerc, G., Dirren, E. & Carrera, E. Machine learning for early dynamic prediction of functional outcome after stroke. *Commun. Med. (Lond.)* **4**, 232 (2024).

6. Paulus, J. K. *et al.* Field synopsis of the role of sex in stroke prediction models. *J. Am. Heart Assoc.* **5**, (2016).

7. Phan, T. G., Clissold, B. B., Ma, H., Van Ly, J. & Srikanth, V. Predicting disability after ischemic stroke based on comorbidity index and stroke severity-from the Virtual International Stroke Trials Archive-acute collaboration. *Front. Neurol.* **8**, 192 (2017).

8. Bonkhoff, A. K. *et al.* Outcome after acute ischemic stroke is linked to sex-specific lesion patterns. *Nat. Commun.* **12**, 3289 (2021).

9. He, Q. *et al.* Predicting a favorable (mRS 0-2) or unfavorable (mRS 3-6) stroke outcome by arterial spin labeling and amide proton transfer imaging in post-thrombolysis stroke patients. *J. Pers. Med.* **13**, 248 (2023).

10. Zhang, Q., Yang, Y. & Saver, J. L. Discharge destination after acute hospitalization strongly predicts three month disability outcome in ischemic stroke. *Restor. Neurol. Neurosci.* **33**, 771–775 (2015).

11. ElHabr, A. K. *et al.* Predicting 90-day modified Rankin Scale score with discharge information in acute ischaemic stroke patients following treatment. *BMJ Neurol. Open* **3**, e000177 (2021).

12. Qureshi, A. I., Chaudhry, S. A., Sapkota, B. L., Rodriguez, G. J. & Suri, M. F. K. Discharge destination as a surrogate for Modified Rankin Scale defined outcomes at 3- and 12-months poststroke among stroke survivors. *Arch. Phys. Med. Rehabil.* **93**, 1408–1413.e1 (2012).

13. Capurro, D., Yetisgen, M., van Eaton, E., Black, R. & Tarczy-Hornoch, P. Availability of structured and unstructured clinical data for comparative effectiveness research and quality improvement: a multisite assessment. *EGEMS (Wash., DC)* **2**, 1079 (2014).

14. Jiang, L. Y. *et al.* Health system-scale language models are all-purpose prediction engines. *Nature* **619**, 357–362 (2023).

15. Thompson, M. P. & Reeves, M. Abstract 168: Assessing the utility of the modified Rankin Scale (mRS) at discharge to predict day 90 outcomes in acute stroke registries. *Circ. Cardiovasc. Qual. Outcomes* **5**, (2012).

16. Ovbiagele, B. & Saver, J. L. Day-90 acute ischemic stroke outcomes can be derived from early functional activity level. *Cerebrovasc. Dis.* **29**, 50–56 (2010).

17. Ovbiagele, B., Lyden, P. D., Saver, J. L. & VISTA Collaborators. Disability status at 1 month is a reliable proxy for final ischemic stroke outcome. *Neurology* **75**, 688–692 (2010).

18. Sung, S.-F., Hsieh, C.-Y. & Hu, Y.-H. Early prediction of functional outcomes after acute

ischemic stroke using unstructured clinical text: Retrospective cohort study. *JMIR Med. Inform.* **10**, e29806 (2022).

19. Sung, S.-F., Chen, C.-H., Pan, R.-C., Hu, Y.-H. & Jeng, J.-S. Natural language processing enhances prediction of functional outcome after acute ischemic stroke. *J. Am. Heart Assoc.* **10**, e023486 (2021).
20. Guo, Y. *et al.* Novel survival features generated by clinical text information and radiomics features may improve the prediction of ischemic stroke outcome. *Diagnostics (Basel)* **12**, 1664 (2022).
21. Heo, J. *et al.* Machine learning-based model for prediction of outcomes in acute stroke. *Stroke* **50**, 1263–1265 (2019).
22. Su, Y.-H. & Tsai, C.-F. Predicting functional outcomes after a stroke event by clinical text notes: A comparative study of traditional machine learning and deep learning methods. *Health Informatics J.* **31**, 14604582251381194 (2025).
23. Jiang, L. Y. *et al.* Generalist foundation models are not clinical enough for hospital operations. *arXiv [cs.CL]* (2025) doi:10.48550/arXiv.2511.13703.
24. Devlin, J., Chang, M.-W., Lee, K. & Toutanova, K. BERT: Pre-training of deep bidirectional Transformers for language understanding. *arXiv [cs.CL]* (2018).
25. Brown, T. B. *et al.* Language Models are Few-Shot Learners. *arXiv [cs.CL]* (2020).
26. Fonarow, G. C. *et al.* Characteristics, performance measures, and in-hospital outcomes of the first one million stroke and transient ischemic attack admissions in get with the guidelines-stroke. *Circ. Cardiovasc. Qual. Outcomes* **3**, 291–302 (2010).
27. Xian, Y. *et al.* Twenty years of sustained improvement in quality of care and outcomes for patients hospitalized with stroke or transient ischemic attack: Data from The Get With The Guidelines-stroke program. *Stroke* **55**, 2599–2610 (2024).

28. Grattafiori, A. *et al.* The Llama 3 herd of models. *arXiv [cs.AI]* (2024).
29. Sellergren, A. *et al.* MedGemma Technical Report. *arXiv [cs.AI]* (2025).
30. Gallifant, J. *et al.* The TRIPOD-LLM reporting guideline for studies using large language models. *Nat. Med.* **31**, 60–69 (2025).
31. Reeves, M. J. *et al.* Twenty years of Get With The Guidelines-stroke: Celebrating past successes, lessons learned, and future challenges. *Stroke* **55**, 1689–1698 (2024).
32. Han, X. *et al.* Neuro Data Hub: A new approach for streamlining medical clinical research. *Neurosurg. Pract.* **6**, e000162 (2025).
33. Hu, E. J. *et al.* LoRA: Low-Rank Adaptation of large language models. *arXiv [cs.CL]* (2021).
34. Samak, Z. A., Clatworthy, P. & Mirmehdi, M. Automatic prediction of stroke treatment outcomes: latest advances and perspectives. *Biomed. Eng. Lett.* **15**, 467–488 (2025).
35. Silva, L. *et al.* Assessment of the modified Rankin scale in electronic health records with a fine-tuned large language model. *medRxiv* (2025) doi:10.1101/2025.04.30.25326777.
36. Heo, T. S. *et al.* Prediction of stroke outcome using natural language processing-based machine learning of radiology report of brain MRI. *J. Pers. Med.* **10**, 286 (2020).
37. Stinear, C. M., Smith, M.-C. & Byblow, W. D. Prediction tools for stroke rehabilitation. *Stroke* **50**, 3314–3322 (2019).
38. Finally, a Replacement for BERT. https://huggingface.co/blog/modernbert.
39. BigBird. https://huggingface.co/docs/transformers/en/model_doc/big_bird.
40. Vishwanath, K. *et al.* Generalist large language models outperform clinical tools on medical benchmarks. *arXiv [cs.CL]* (2025) doi:10.48550/arXiv.2512.01191.

41. Alyakin, A. *et al.* CNS-Obsidian: A Neurosurgical Vision-Language Model Built From Scientific Publications. *arXiv [cs.AI]* (2025).

42. Wilson, J. T. L. *et al.* Reliability of the modified Rankin Scale across multiple raters: benefits of a structured interview: Benefits of a structured interview. *Stroke* **36**, 777–781 (2005).

43. Erdoğan, M. Ş. *et al.* Predicting modified Rankin scale scores of ischemic stroke patients using radiomics features and machine learning. in *Advances in Intelligent Systems and Computing* 204–213 (Springer Nature Switzerland, Cham, 2024).

44. Alyakin, A. *et al.* CNS-CLIP: Transforming a neurosurgical journal into a multimodal medical model. *Neurosurgery* (2024) doi:10.1227/neu.0000000000003297.

45. Kondepudi, A. *et al.* Health system learning achieves generalist neuroimaging models. *arXiv [cs.CV]* (2025) doi:10.48550/arXiv.2511.18640.

46. Pedro, T. *et al.* Exploring the use of ChatGPT in predicting anterior circulation stroke functional outcomes after mechanical thrombectomy: a pilot study. *J. Neurointerv. Surg.* **17**, 261–265 (2025).

47. Chen, S. F. *et al.* LLM-assisted systematic review of large language models in clinical medicine. *Nat. Med.* (2026).

48. Singh, S. *et al.* The pitfalls of multiple-choice questions in generative AI and medical education. *Sci. Rep.* **15**, 42096 (2025).

49. Vishwanath, K. *et al.* Medical large language models are easily distracted. *arXiv [cs.CL]* (2025).

**Supplemental Figures**

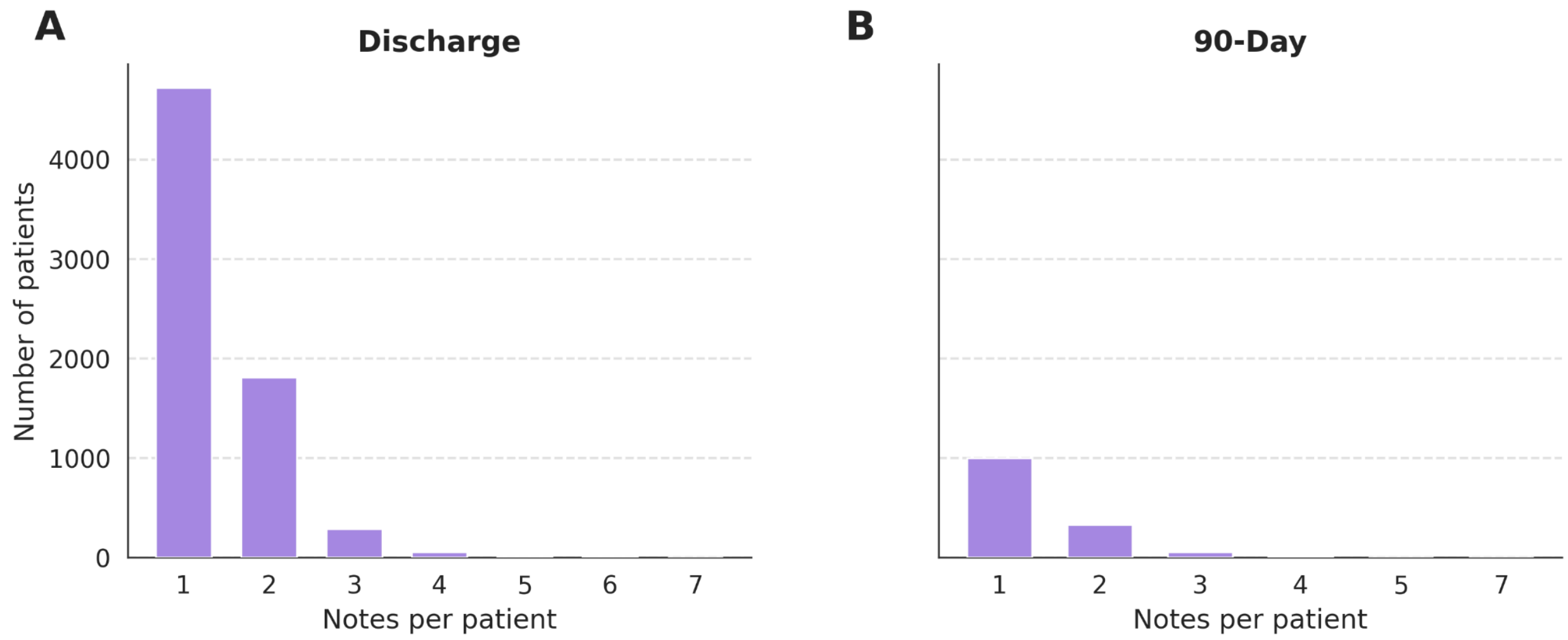

**Figure S1. Distribution of H&P notes per patient at discharge and 90-day follow-up.** Panels A and B summarize the number of notes per patient in the discharge cohort (A) and the subset with available 90-day follow-up (B).

Abbreviations: *H&P,* History and Physical

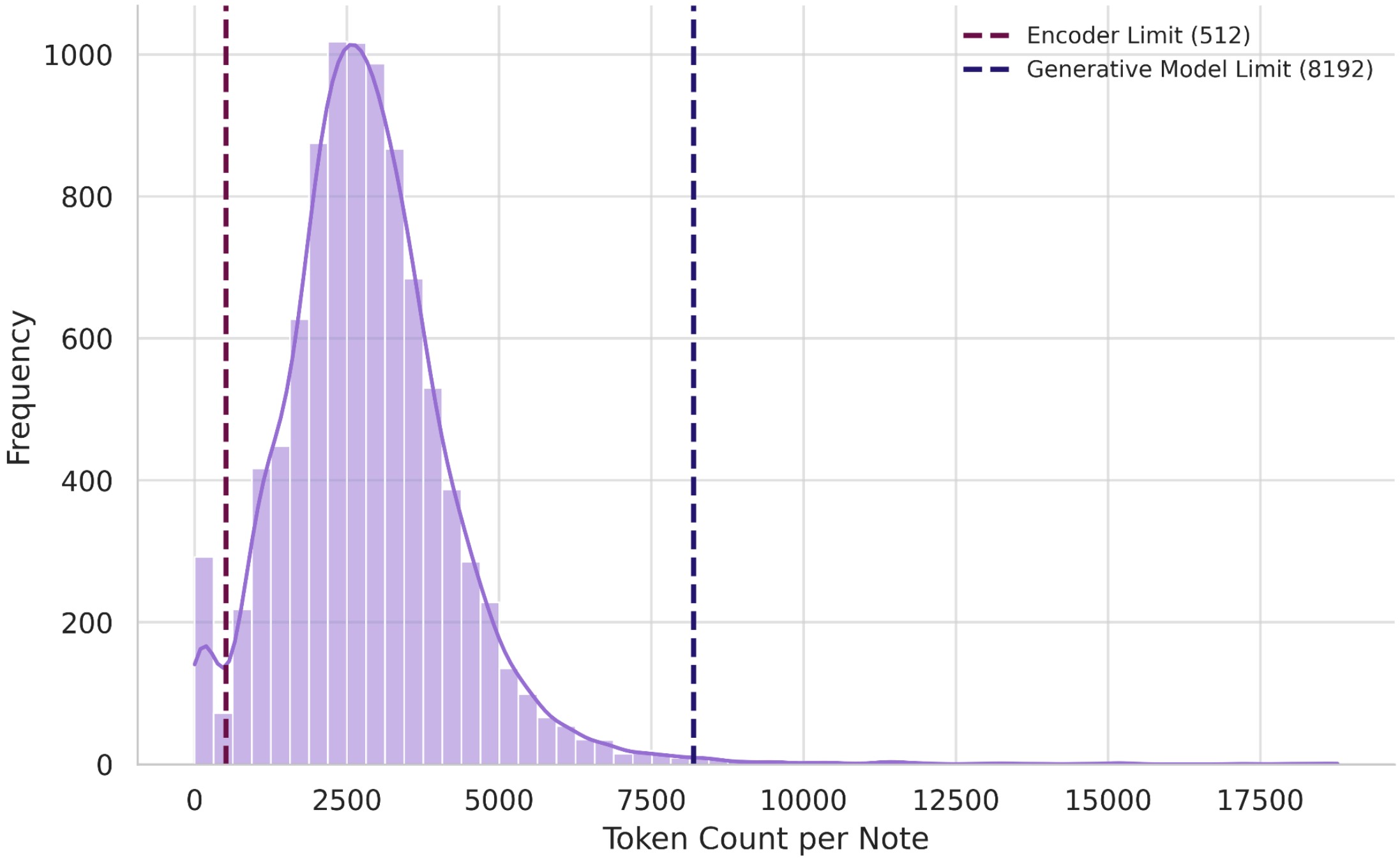

**Figure S2. Distribution of H&P token lengths across all admission notes.** Token lengths were computed using the BERT WordPiece tokenizer. Dashed vertical lines indicate the 512-token limit of encoder models and the 8192-token context window typical of generative LLMs.

Abbreviations: *H&P,* History and Physical

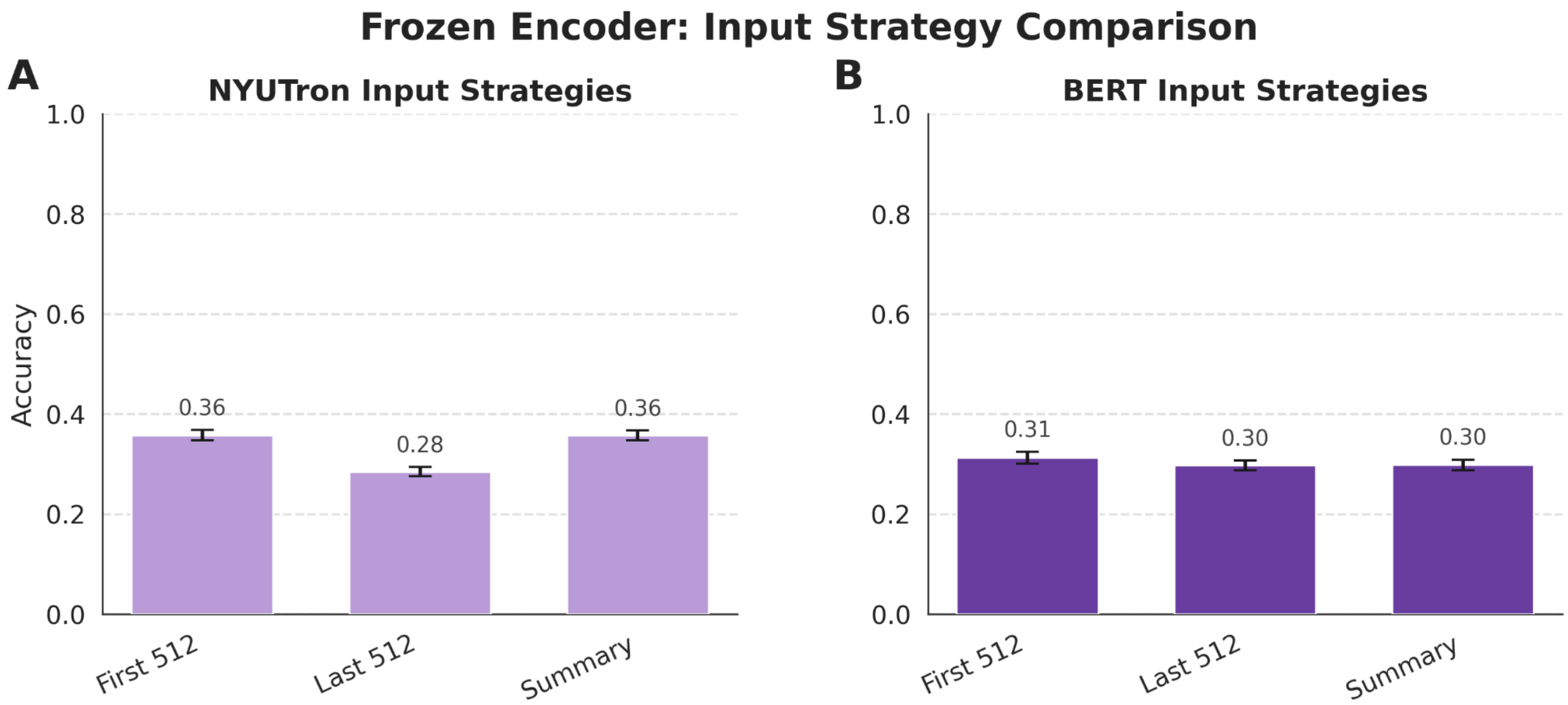

**Figure S3. Comparison of input strategies for NYUTron and BERT encoders.** Panels A–B show mean accuracy for frozen NYUTron (Panel A) and frozen BERT (Panel B) embeddings derived using three input strategies: the first 512 tokens of the admission H&P note, the last 512 tokens, and a Llama-generated clinical summary. All models use a random forest classifier trained on frozen encoder embeddings. Error bars indicate 95% confidence intervals obtained via 10-fold cross-validation on the training set.

Abbreviations**:** *BERT*, Bidirectional Encoder Representations from Transformers; *H&P*, History and Physical; *LLM*, large language model.

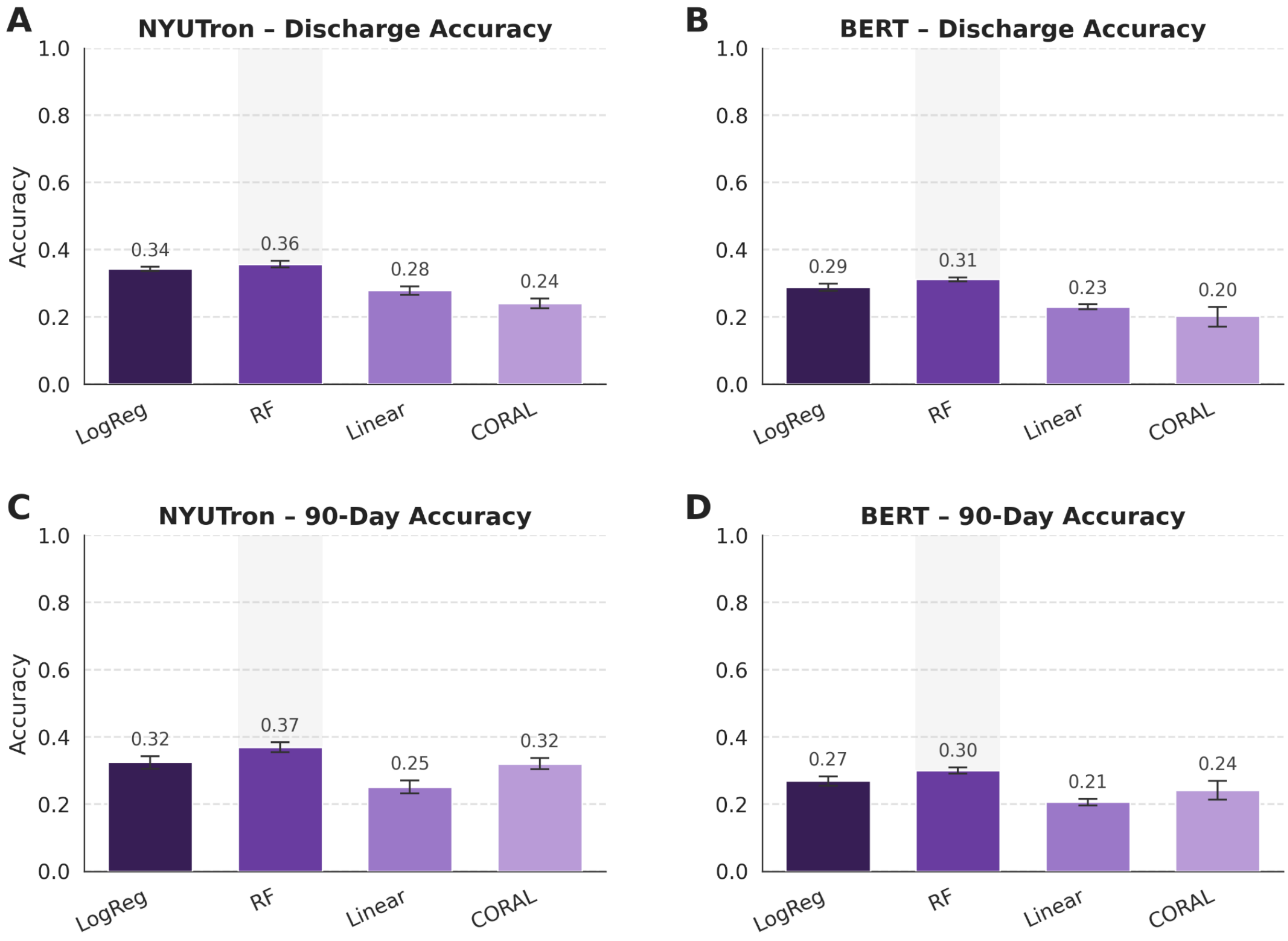

**Figure S4. Comparison of classifier heads for NYUTron and BERT encoders.** Panels A–D show mean accuracy for four downstream classifier heads—logistic regression (LogReg), random forest (RF; 500 trees, max depth = 7), linear regression (Linear; predictions rounded and clipped to the valid mRS range 0–6), and CORAL ordinal regression—trained on frozen embeddings extracted from admission H&P notes. Panels A–B present discharge accuracy for NYUTron (Panel A) and BERT (Panel B), while Panels C–D present 90-day accuracy for NYUTron (Panel C) and BERT (Panel D). Error bars indicate 95% confidence intervals obtained via 10-fold cross-validation on the training set.

Abbreviations: *BERT*, Bidirectional Encoder Representations from Transformers; *CORAL*, COntinuous RAnked Logits; *H&P,* History and Physical; mRS, modified Rankin Scale; *LogReg,* Logistic Regression; *RF,* Random Forest

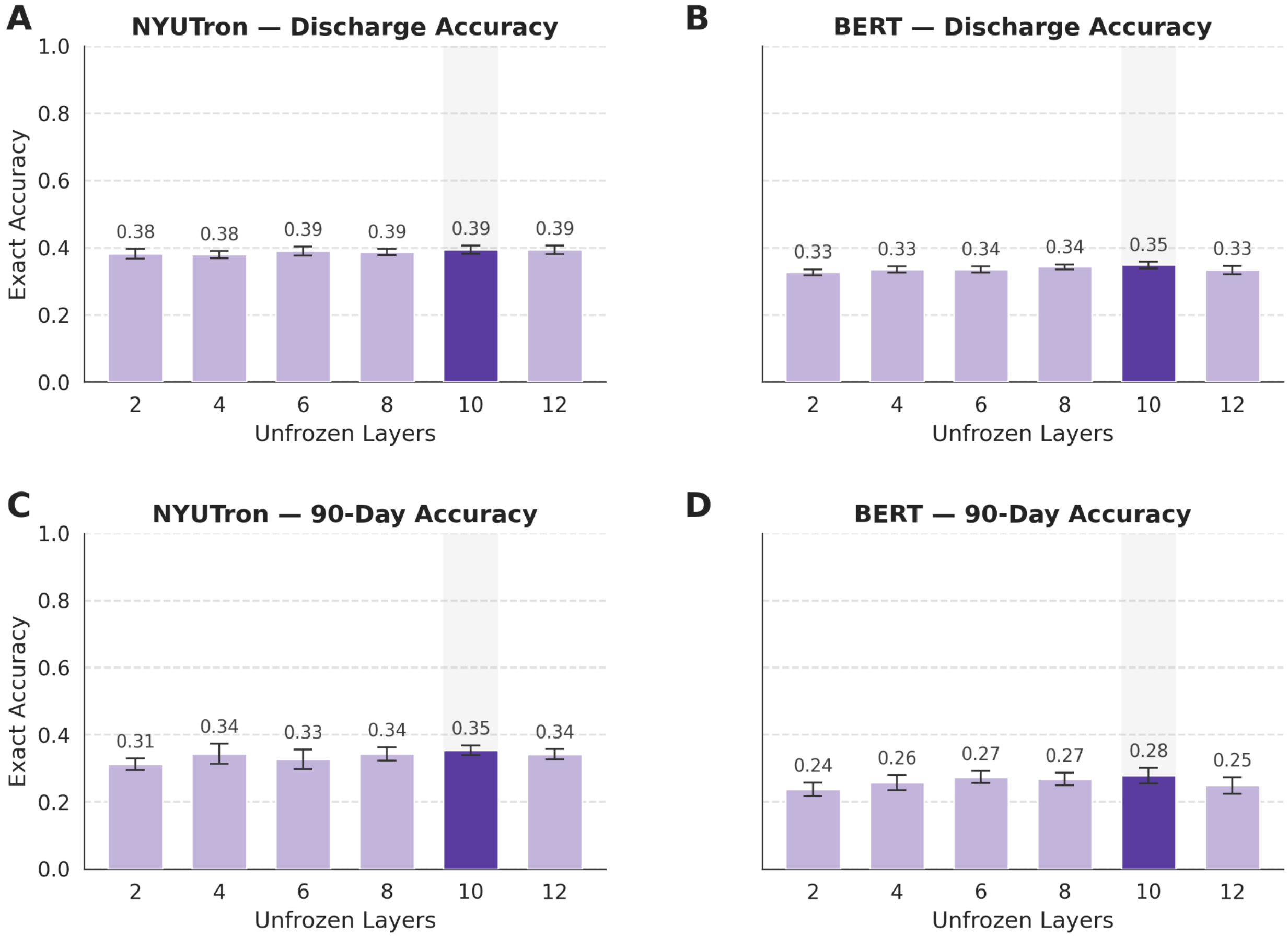

**Figure S5. Fine-tuning depth comparison for NYUTron and BERT encoders.** Panels A–D show mean accuracy for NYUTron (Panels A, C) and BERT (Panels B, D) when fine-tuned to different depths. Models were fine-tuned by unfreezing between 2 and 12 transformer layers. For each depth, the encoder's [CLS] representation was passed through a jointly trained linear regression head to predict continuous mRS scores (0–6), optimized using MAE loss. Depth 10—identified as the optimal fine-tuning depth across both encoders and timepoints—is highlighted in dark purple. Panels A–B report discharge accuracy for NYUTron and BERT, respectively, and Panels C–D report 90-day accuracy. Error bars indicate 95% confidence intervals obtained via 10-fold cross-validation on the training set.

Abbreviations: *BERT*, Bidirectional Encoder Representations from Transformers; *CLS*, classification token; *MAE*, mean absolute error; *mRS*, modified Rankin Scale

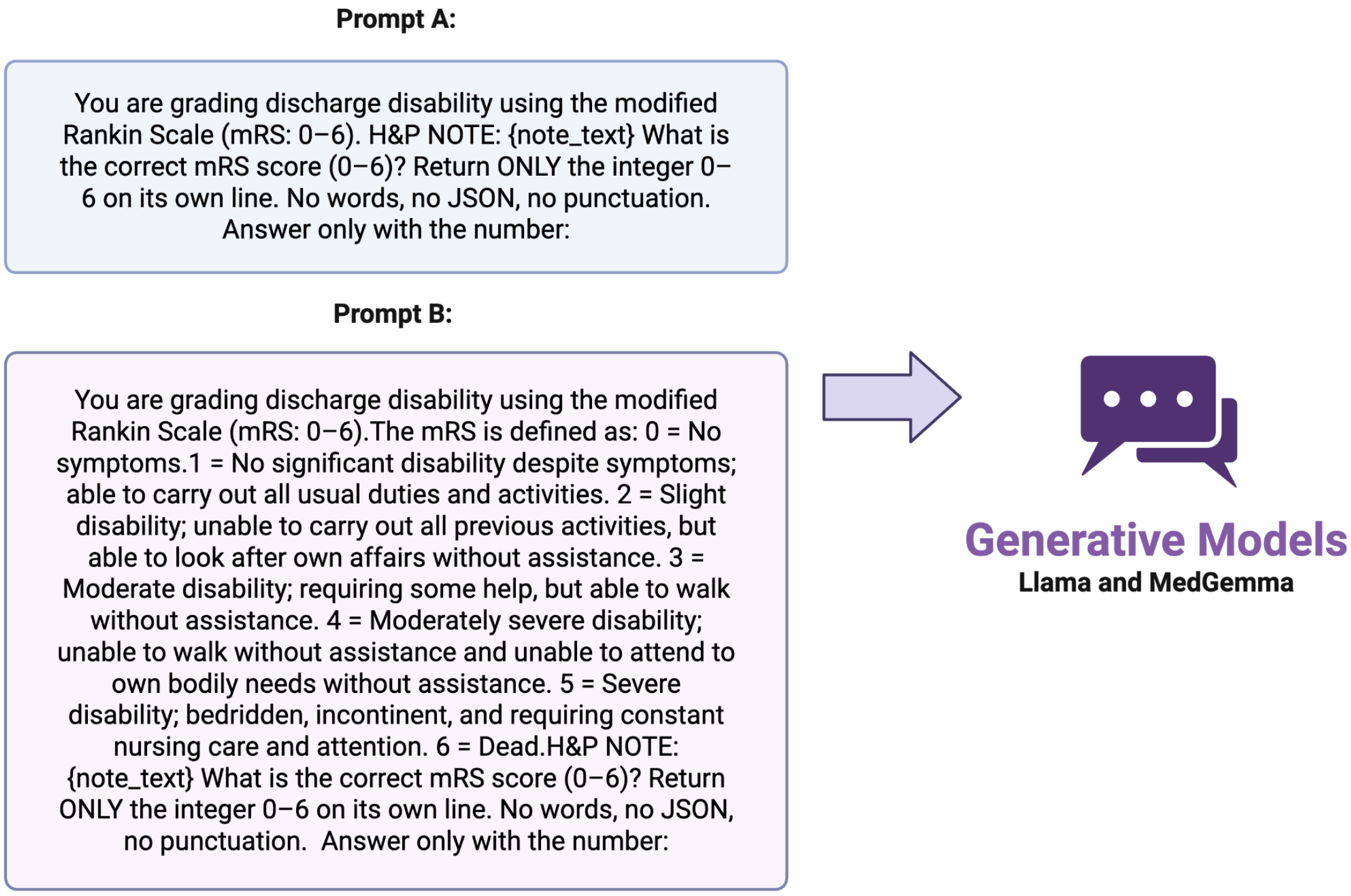

**Figure S6. Prompting strategy comparison for zero-shot mRS prediction.** Two prompting strategies were evaluated for zero-shot discharge mRS prediction using generative models (Llama and MedGemma). Prompt A provides a concise instruction asking the model to output a single integer mRS score (0–6) based solely on the H&P note. Prompt B augments this instruction by including the full mRS definitions (0–6) before requesting the prediction. Both prompts explicitly require the model to return only the integer label with no additional text. As there was no significant difference between prompts, Prompt B was used.

Abbreviations: *H&P,* History and Physical; *mRS*, modified Rankin Scale

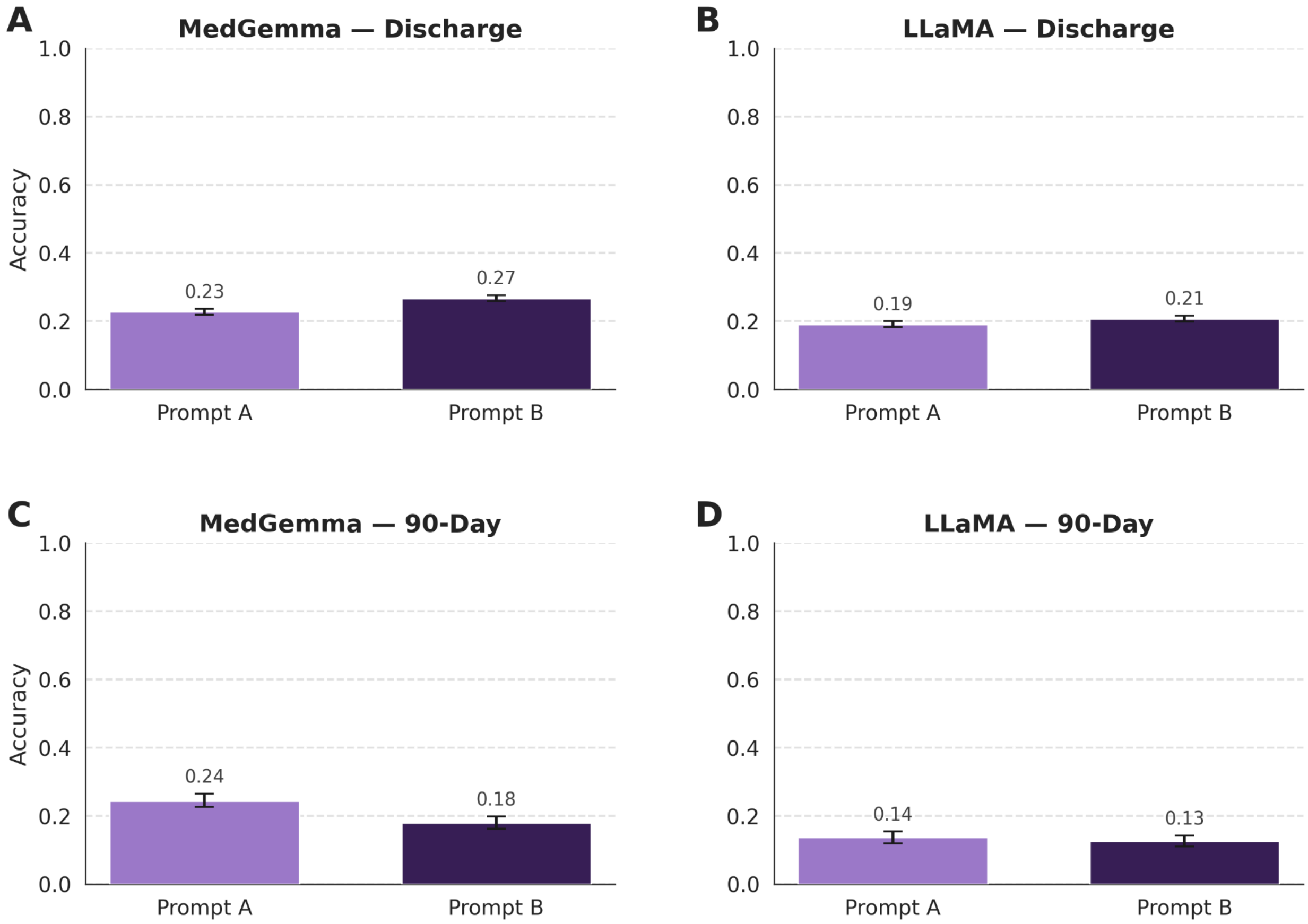

**Figure S7. Comparison of zero-shot prompt designs for MedGemma and Llama.** Panels A–D show zero-shot accuracy for MedGemma (Panels A and C) and Llama (Panels B and D) on discharge (Panels A–B) and 90-day (Panels C–D) mRS prediction using two prompt variants. Prompt A instructed the model only to predict the mRS score, whereas Prompt B additionally provided the mRS category definitions. Error bars indicate 95% confidence intervals obtained via bootstrapping. Results throughout the study are reported using Prompt B.

Abbreviations: *mRS*, modified Rankin Scale

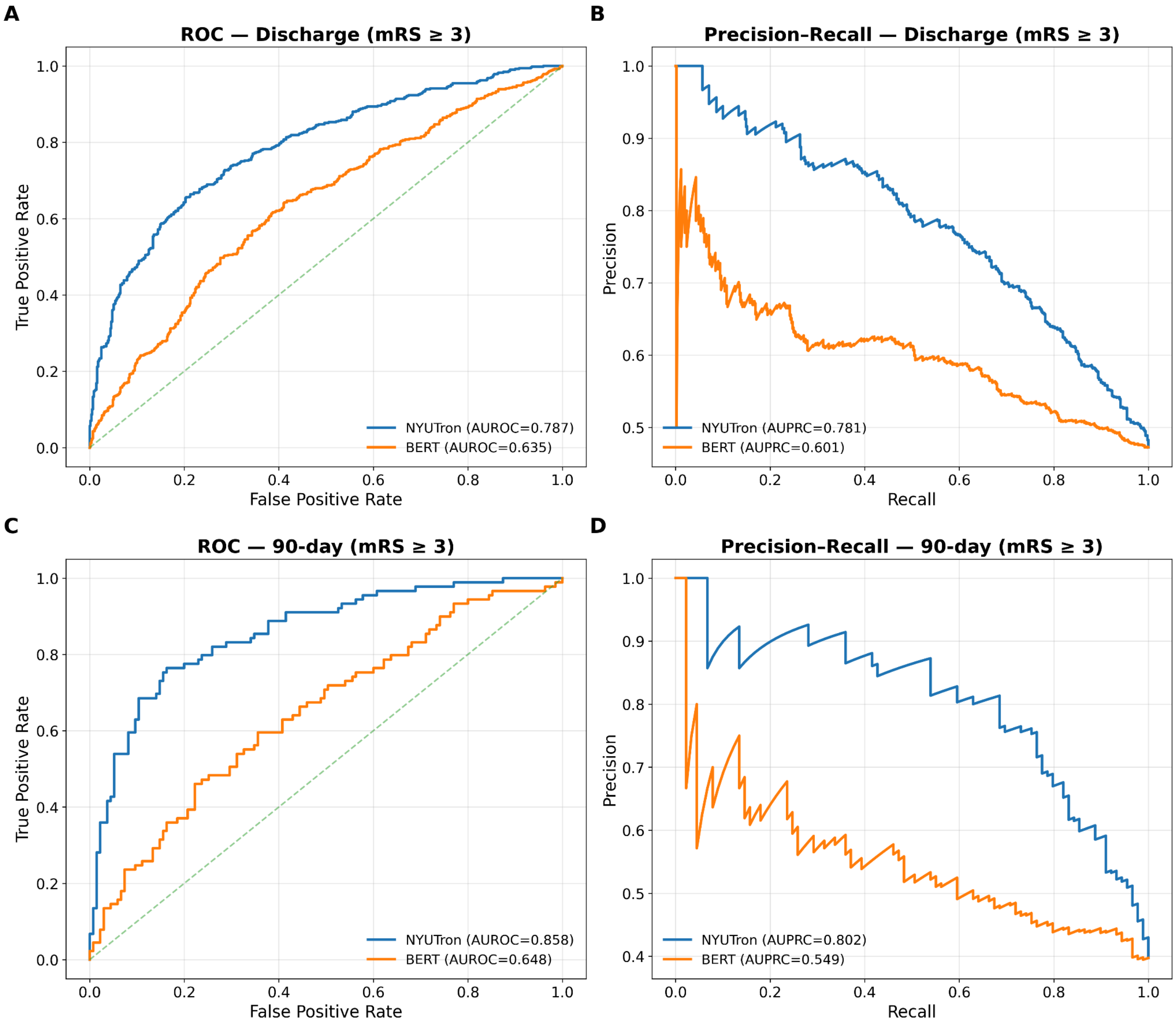

**Figure S8. Discrimination performance of frozen encoder models for binary mRS prediction.** Panels A–D show receiver operating characteristic (ROC) curves (Panels A and C) and precision–recall (PR) curves (Panels B and D) for frozen NYUTron and BERT models in predicting unfavorable functional outcome (mRS ≥ 3) at hospital discharge (Panels A–B) and 90 days post-discharge (Panels C–D). Continuous confidence scores for binary mRS ≥ 3 classification were obtained from random forest classifiers trained on frozen encoder embeddings. Legends report area-under-the-curve metrics for each model. Zero-shot prompted decoder models (Llama and MedGemma) were not included, as they produced only discrete mRS predictions without continuous confidence scores, precluding ROC and precision–recall analysis.

Abbreviations: *AUC*, area under the curve; *mRS*, modified Rankin Scale; *PR*, precision–recall; *ROC*, receiver operating characteristic

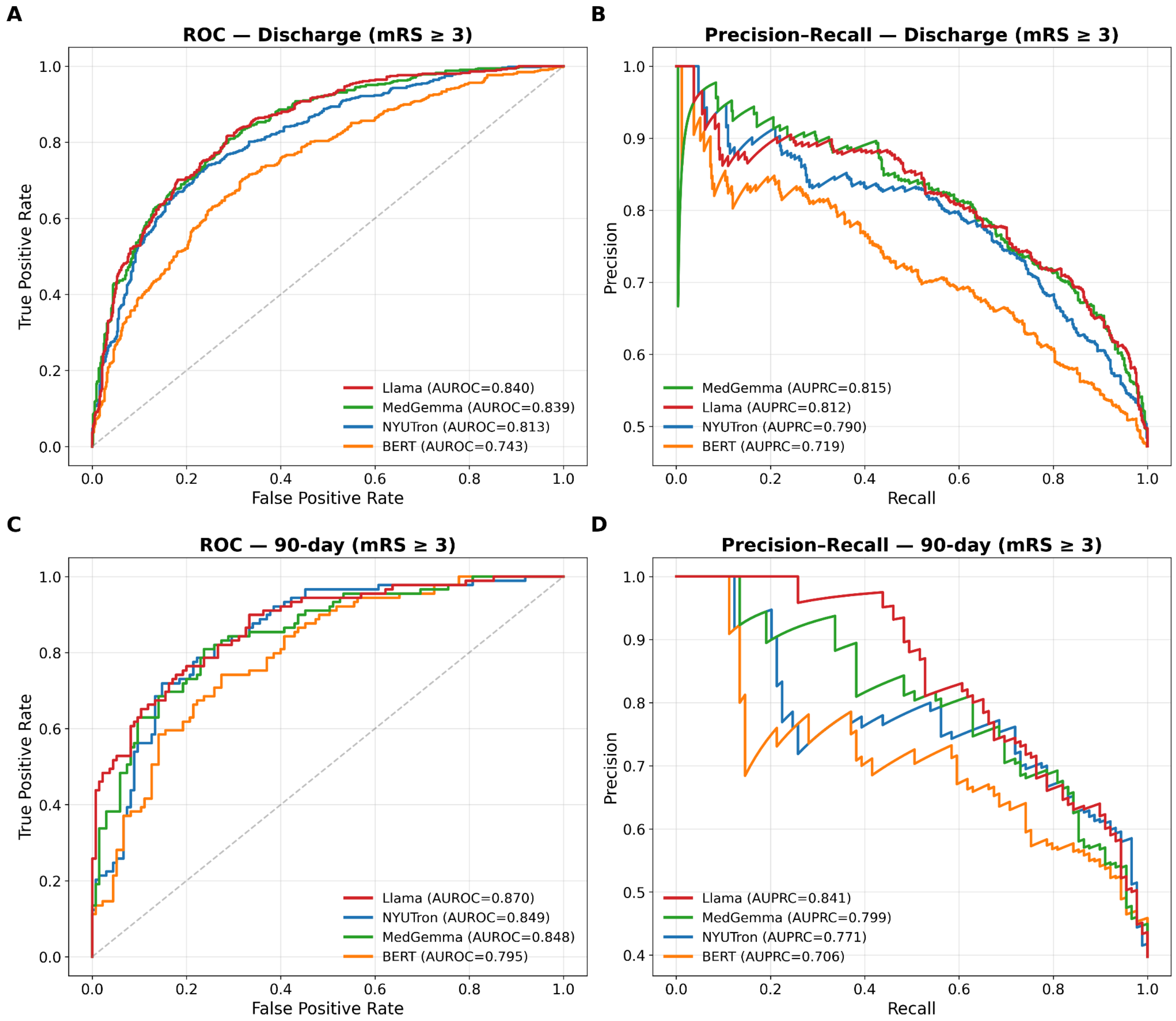

**Figure S9. Discrimination performance of fine-tuned encoder and decoder models for binary mRS prediction.** Panels A–D show receiver operating characteristic (ROC) curves (Panels A and C) and precision–recall (PR) curves (Panels B and D) for fine-tuned NYUTron, BERT, MedGemma, and Llama models in predicting unfavorable functional outcome (mRS ≥ 3) at hospital discharge (Panels A–B) and 90 days post-discharge (Panels C–D). Continuous confidence scores for binary mRS ≥ 3 classification were obtained from fine-tuned model outputs; for encoder models, confidence scores were derived from regression outputs, while decoder model scores were obtained by aggregating predicted class probabilities for mRS ≥ 3. Legends report area-under-the-curve metrics for each model.

Abbreviations: *AUC*, area under the curve; *mRS*, modified Rankin Scale; *PR*, precision–recall; *ROC*, receiver operating characteristic

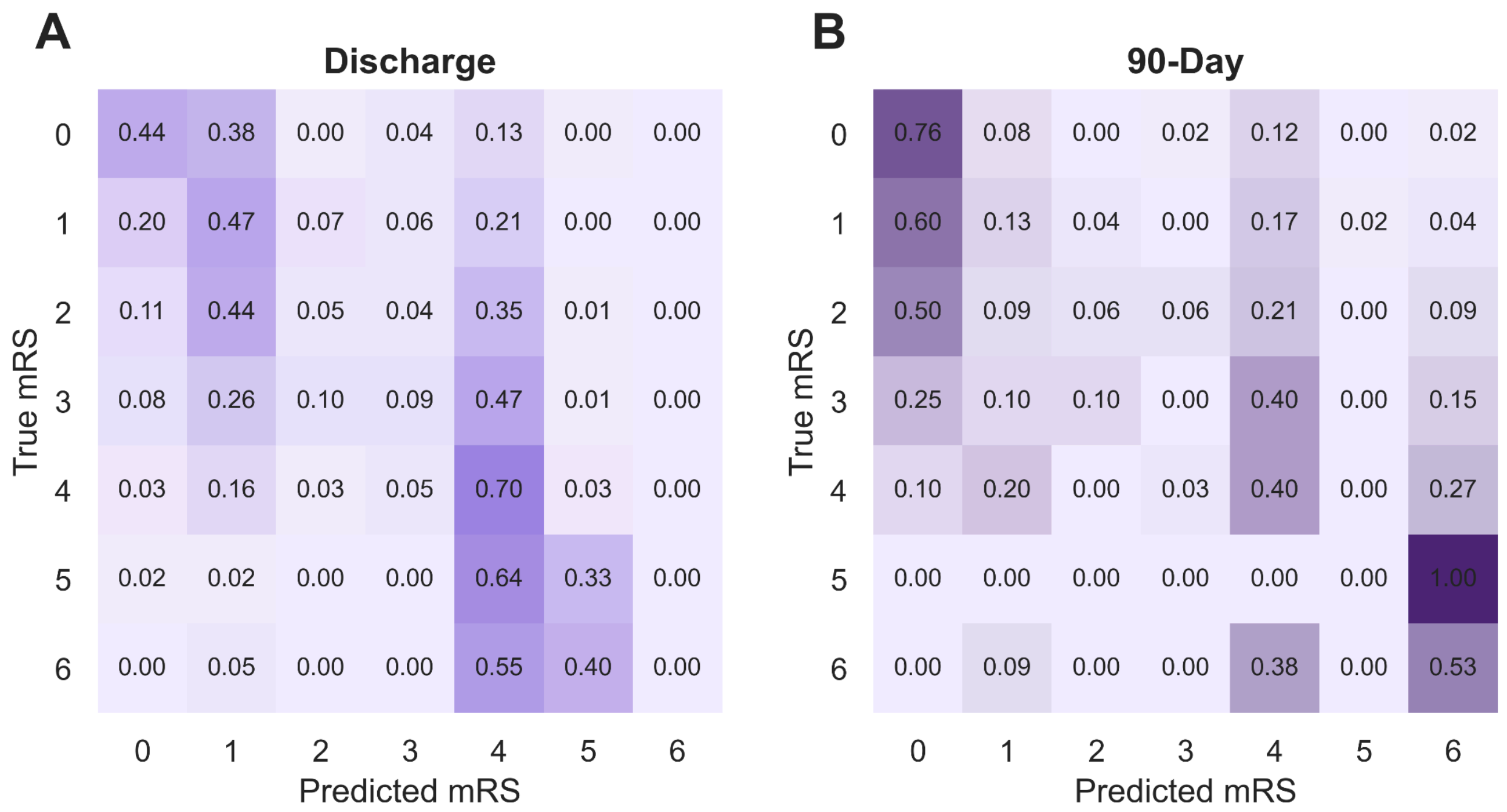

**Figure S10. Normalized confusion matrices for the best-performing model, fine-tuned Llama-3.1-8B, at discharge and 90-day follow-up.** Panels A and B show normalized confusion matrices for discharge (Panel A) and 90-day (Panel B) mRS prediction. Each matrix displays per-class prediction accuracy for mRS scores (0–6), normalized by the true label distribution.

Abbreviations: *mRS*, modified Rankin Scale